\pdfoutput=1
\documentclass{ieeeaccess}
\usepackage[utf8]{inputenc}
\usepackage{newunicodechar}

\usepackage{graphicx}
\usepackage{amsmath,amssymb,amsfonts,amsthm}
\usepackage{booktabs}
\usepackage{multirow}
\usepackage{array}
\usepackage[hidelinks]{hyperref}
\usepackage{xcolor}
\definecolor{accessblue}{RGB}{0,98,155}
\usepackage{algorithm}
\usepackage{algpseudocode}
\usepackage{subcaption}
\usepackage{cite}
\usepackage{url}
\usepackage{microtype}
\usepackage{enumitem}
\usepackage{caption}
\usepackage{float}
\usepackage{mathtools}
\usepackage{bm}
\makeatletter
\let\access@cls@year\year
\def\year{\pdfprimitive\year}
\makeatother
\usepackage{tikz}
\usepackage{pgfplots}
\pgfplotsset{compat=1.18}
\usetikzlibrary{shapes.geometric,arrows.meta,positioning,calc,fit,backgrounds}
\makeatletter
\let\year\access@cls@year
\makeatother
\def\theyear{2026}
\def\thevol{14}
\usepackage{rotating}
\usepackage{tabularx}
\usepackage{makecell}
\usepackage{pifont}
\usepackage{etoolbox}

\AtBeginEnvironment{thebibliography}{\raggedright}

\newunicodechar{≈}{\ensuremath{\approx}}
\newunicodechar{×}{\ensuremath{\times}}
\newunicodechar{≥}{\ensuremath{\geq}}
\newunicodechar{≤}{\ensuremath{\leq}}
\newunicodechar{→}{\ensuremath{\rightarrow}}
\newunicodechar{−}{\ensuremath{-}}
\newunicodechar{—}{---}
\newunicodechar{–}{--}

\newcommand{\safeincludegraphics}[2][]{%
  \IfFileExists{#2}{\includegraphics[#1]{#2}}{%
    \fbox{\parbox[c][4cm][c]{0.9\linewidth}{\centering Missing figure: \\ \texttt{#2}}}%
  }%
}

\newcommand{\vmem}{V_{\mathrm{mem}}}
\newcommand{\vth}{V_{\mathrm{th}}}
\newcommand{\vleak}{V_{\mathrm{leak}}}
\newcommand{\vtau}{\tau}
\newcommand{\vte}{\textit{v2e}}
\newcommand{\csnn}{\textsc{Conv-SNN}}

\newcommand{\jaad}{\textsc{JAAD}}

\newcommand{\dvspedx}{\textsc{DVS-PedX}}
\newcommand{\real}[1]{\textbf{#1}}
\newcommand{\cmark}{\ding{51}}
\newcommand{\xmark}{\ding{55}}

\begin{document}
\history{Date of publication xxxx 00, 0000, date of current version xxxx 00, 0000.}
\doi{10.1109/ACCESS.2026.Doi Number}

\title{Pedestrian Crossing Intent Classification From Event-Based
Vision Using Convolutional Spiking Neural Networks With
Temporal Augmentation}

\author{%
\uppercase{Henok Teklu}\authorrefmark{1},
\uppercase{Mustafa Sakhai}\authorrefmark{2},
\uppercase{Maciej Wielgosz}\authorrefmark{2},
\and \uppercase{Matej Mertik}\authorrefmark{1}}

\address[1]{Applied Artificial Intelligence, Alma Mater Europaea University, Maribor, Slovenia (e-mail: henok.teklu@almamater.si; matej.mertik@almamater.si)}
\address[2]{AGH University of Krak\'{o}w, Krak\'{o}w, Poland (e-mail: msakhai@agh.edu.pl; wielgosz@agh.edu.pl)}
\tfootnote{This work was carried out within the PhD programme in Applied
Artificial Intelligence at Alma Mater Europaea University, in collaboration
with AGH University of Krak\'{o}w.}

\markboth
{Teklu \headeretal: Pedestrian Crossing Intent Classification From Event-Based Vision Using Convolutional SNNs}
{Teklu \headeretal: Pedestrian Crossing Intent Classification From Event-Based Vision Using Convolutional SNNs}

\corresp{Corresponding authors: Henok Teklu (e-mail: henok.teklu@almamater.si)
and Mustafa Sakhai (e-mail: msakhai@agh.edu.pl).}

\begin{abstract}
Anticipating whether a pedestrian will cross the road is safety-critical for
autonomous vehicles, requiring real-time inference under challenging conditions
including motion blur, high dynamic range, and class imbalance.  Conventional
frame-based deep networks process redundant RGB data at fixed frame rates,
limiting their temporal resolution and energy efficiency.  In this work we
present an end-to-end pipeline that (i) converts real-world driving footage
from the Joint Attention in Autonomous Driving (JAAD) dataset into synthetic
dynamic vision sensor (DVS)
event streams using the v2e simulator, (ii) augments training with the
CARLA-simulated DVS sequences of the DVS-PedX dataset under both normal and
adverse weather conditions, and (iii)
trains a novel convolutional spiking neural network (Conv-SNN) with
clip-consistent DVS augmentation to classify pedestrian crossing intent as
binary: crossing or non-crossing.  We detail all architectural
decisions, the exact leaky-integrate-and-fire neuron dynamics with
surrogate-gradient learning, the
class-balanced loss formulation, JAAD oversampling at $6\times$, and a
70/15/15 stratified splitting protocol.  The trained model achieves
95.83\% accuracy and F1\,=\,0.9695 on the JAAD DVS test set,
97.79\% accuracy and F1\,=\,0.9478 on normal CARLA DVS, and
94.78\% accuracy and F1\,=\,0.8369 on adverse-weather CARLA
DVS---all from a 1.07\,M-parameter architecture trained on CPU.  Compared to
prior frame-based approaches on JAAD, our method closes or surpasses the
reported accuracy while operating natively on sparse temporal representations.
We include a thorough analysis of the convergence behaviour across all 15
training epochs, domain transfer characteristics, and a quantitative comparison
with representative related work.
\end{abstract}


\titlepgskip=-21pt

\maketitle

\section{Introduction}
\label{sec:intro}

\PARstart{A}{utonomous} driving safety relies critically on early, accurate prediction of
pedestrian intent.  Unlike obstacle detection---where the object is already in
the scene---intent prediction must infer a future action from subtle pre-movement
cues: postural shifts, head orientation, gaze direction, and micro-gestures
that unfold in tens to hundreds of milliseconds before any foot leaves the
pavement.  Failure to detect a crossing pedestrian in time can be fatal; false
positives impose unnecessary braking, degrading ride comfort and traffic flow.

Standard approaches model this as a sequence classification problem using
red-green-blue (RGB) video from vehicle-mounted cameras~\cite{rasouli2019jaad,kotseruba2021benchmark}.
The success of deep convolutional neural networks (CNNs)~\cite{lecun1998gradient,krizhevsky2012alexnet,he2016resnet}
and recurrent architectures~\cite{hochreiter1997lstm} in video understanding
has driven CNN-LSTM (long short-term memory) and Transformer-based architectures to achieve strong accuracy on
curated benchmark datasets~\cite{rasouli2019jaad,pedformer2023,goodfellow2016deep}.  However, they
share a fundamental limitation: frame-rate cameras sample the scene uniformly
at 25--60\,fps regardless of scene dynamics.  Slow pedestrians generate mostly
redundant frames; fast or close pedestrians saturate the sensor and introduce
motion blur.  High-dynamic-range scenarios (tunnel exits, night driving with
headlights) cause further degradation.  Processing full RGB frames at inference
time also demands substantial compute, limiting deployment on power-constrained
embedded platforms.

\begin{figure}[tbp]
\centering
\resizebox{\columnwidth}{!}{%
\begin{tikzpicture}[font=\scriptsize, >={Stealth[length=1.8mm]}]
  \begin{scope}
    \foreach \y in {0.4,0.8,1.2,1.6}
      \draw[gray!35, dashed] (0,\y) -- (8.6,\y);
    \draw[->, black!70] (0,0) -- (8.85,0) node[below right=-1pt and -3pt]{time $t$};
    \draw[->, black!70] (0,0) -- (0,2.05) node[left]{$\log I$};
    \draw[blue!70!black, line width=1pt]
      plot[smooth, tension=0.7] coordinates
      {(0,0.25)(0.8,0.35)(1.6,0.7)(2.4,1.15)(3.2,1.55)(4.0,1.78)
       (4.8,1.55)(5.6,1.2)(6.4,0.85)(7.2,0.55)(8.0,0.45)};
    \foreach \p in {(1.25,0.4),(1.95,0.8),(2.75,1.2),(3.55,1.6),
                    (5.05,1.6),(5.8,1.2),(6.6,0.8),(7.6,0.4)}
      \fill[black] \p circle (1.1pt);
    \node[blue!70!black, anchor=east] at (8.5,1.93) {$\log I(t)$};
    \node[gray!55!black, anchor=west] at (0.1,1.9) {threshold steps $\theta$};
  \end{scope}
  \begin{scope}[yshift=-1.75cm]
    \draw[->, black!70] (0,0) -- (8.85,0) node[below right=-1pt and -3pt]{time $t$};
    \draw[black!20] (0,0.55) -- (8.6,0.55);
    \draw[black!20] (0,-0.55) -- (8.6,-0.55);
    \node[anchor=east, green!50!black] at (-0.08,0.55) {ON};
    \node[anchor=east, red!75!black]  at (-0.08,-0.55) {OFF};
    \foreach \x in {1.25,1.95,2.75,3.55}
      \draw[green!55!black, line width=1pt, ->] (\x,0) -- (\x,0.55);
    \foreach \x in {5.05,5.8,6.6,7.6}
      \draw[red!75!black, line width=1pt, ->] (\x,0) -- (\x,-0.55);
  \end{scope}
\end{tikzpicture}%
}
\caption{Operating principle of a Dynamic Vision Sensor.  Each pixel tracks its
  log-intensity $\log I(t)$ (top) and emits an asynchronous \emph{event}
  whenever the signal changes by a fixed contrast quantum $\theta$ (dashed
  levels, marked $\bullet$): an \textcolor{green!55!black}{ON} event on an
  increase and an \textcolor{red!75!black}{OFF} event on a decrease (bottom).
  Static regions emit no events, yielding the sparse, microsecond-resolution
  stream our \csnn{} consumes.}
\label{fig:dvs_principle}
\end{figure}
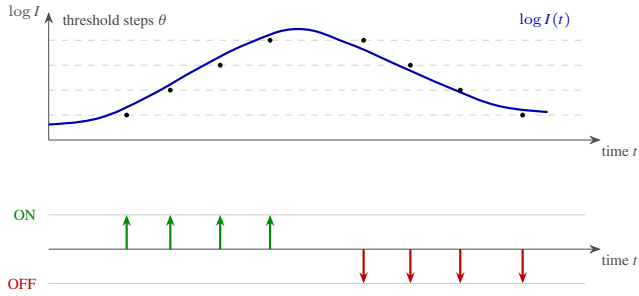

\paragraph{Event-based cameras.}
Dynamic Vision Sensors (DVS)~\cite{lichtsteiner2008dvs,gallego2020eventsurvey,indiveri2011neuromorphic}
overcome these limitations by detecting per-pixel log-luminance changes
asynchronously.  Each pixel fires an \emph{event} $(x, y, t, p)$ only when
the change $|\Delta \ln I|$ exceeds a threshold $\theta$, yielding a sparse
stream with microsecond temporal resolution, $>$120\,dB dynamic range, and
typical power draw of 5--20\,mW.  Redundant background pixels produce no
events; moving objects produce dense, time-stamped event clouds.  This makes
DVS cameras inherently well-suited to motion-based intent prediction tasks.
Figure~\ref{fig:dvs_principle} illustrates this event-generation principle.

\paragraph{Spiking Neural Networks.}
Spiking Neural Networks (SNN)~\cite{maass1997snns,burkitt2006lif} process information using
binary spike events rather than continuous activations, mirroring the
communication mechanism of biological neurons~\cite{dayan2001theoretical,hubel1968receptive}.
Leaky-Integrate-and-Fire (LIF)
neurons accumulate input over time and fire when a threshold is crossed,
naturally integrating temporal information within the membrane potential.  When
implemented on neuromorphic hardware (Intel Loihi~\cite{davies2018loihi}, IBM
TrueNorth~\cite{merolla2014truenorth}), SNNs can achieve 10--50$\times$ energy reduction over GPU inference.
Training is enabled by the surrogate gradient technique~\cite{neftci2019surrogate,zenke2018superspike},
which replaces the non-differentiable Heaviside spike function with a smooth
approximation during backpropagation~\cite{wu2018snn_backprop,lee2016training}.

\paragraph{The gap.}
Despite strong progress in pedestrian intent prediction with
CNNs~\cite{kotseruba2021benchmark,pedformer2023,saleh2019realtime} and in event-based action
recognition~\cite{wang2019eventaction,amir2017gesture,li2022eventdet}, the intersection of
DVS-based perception and SNN classifiers for \emph{pedestrian intent} remains
largely unexplored.  Existing crossing prediction methods operate on RGB frames;
existing SNN studies on DVS data focus on synthetic gesture or object datasets
far removed from autonomous driving.

\paragraph{This paper.}
We close this gap with a complete, reproducible pipeline.  Our contributions are:

\begin{enumerate}[leftmargin=1.4em,topsep=2pt,itemsep=2pt]
  \item \textbf{End-to-end synthetic DVS pipeline.}  We convert 346 Joint
    Attention in Autonomous Driving (JAAD)
    pedestrian behaviour clips to DVS event frames via \vte{} using documented
    parameters ($\theta{=}0.15$, $\sigma{=}0.03$, $f_c{=}15$\,Hz), discarding
    empty frames through a principled grey-level test.

  \item \textbf{Multi-source training with principled oversampling.}  JAAD clips
    are combined with the CARLA normal and adverse-weather DVS sequences of the
    \dvspedx{} dataset; JAAD is
    oversampled $6\times$ to match the CARLA volume, with each repeat receiving
    independently sampled augmentations.

  \item \textbf{Clip-consistent DVS augmentation.}  Three augmentation
    operations---horizontal flip, brightness jitter, and random spatial
    crop---are applied identically across all $T{=}9$ frames of a clip to
    preserve temporal coherence.

  \item \textbf{Convolutional LIF architecture with class-balanced training.}
    A three-block convolutional SNN (1.07\,M parameters) with AdamW,
    ReduceLROnPlateau, and inverse-frequency class weights, achieving
    \real{95.83\%}/\real{F1\,0.9695} on JAAD and \real{97.79\%}/\real{F1\,0.9478}
    on CARLA---from a CPU-only training run.
\end{enumerate}

The remainder of the paper is organised as follows.  Section~\ref{sec:related}
surveys related work and positions our approach.  Section~\ref{sec:dvs}
details the DVS generation pipeline.  Section~\ref{sec:method} describes the
full \csnn{} model, loss, and training procedure.  Section~\ref{sec:experiments}
reports experimental setup and all results.  Sections~\ref{sec:analysis},
\ref{sec:discussion}, \ref{sec:limitations}, and~\ref{sec:future} provide
analysis, discussion, limitations, and future directions.

\section{Related Work}
\label{sec:related}

\subsection{Pedestrian Crossing Intention Prediction}

\paragraph{Classical and hybrid methods.}
Early crossing prediction relied on hand-crafted features---pose estimates,
heading vectors, optical flow---fed to support vector machines (SVMs) or
hidden Markov models (HMMs)~\cite{schneemann2016context,kooij2014context,volz2016feature}.  These methods are brittle to occlusion and
generalise poorly across environments.  Rasouli~et~al.~\cite{rasouli2017anticipating}
were among the first to apply deep learning to the problem, using a
CNN-LSTM architecture on dashcam footage.  Their \textit{JAAD} benchmark~\cite{rasouli2019jaad}
provided 346 labelled clips from real European and
North American driving, enabling systematic comparison.
The related \textit{PIE} dataset~\cite{rasouli2020pie} further expanded the scope
with ego-motion and traffic signals.  Fang and L\'{o}pez~\cite{fang2018pedestrian}
demonstrated the value of skeleton pose features for crossing intent,
while Gesnouin~et~al.~\cite{gesnouin2020predicting} and Kwak~et~al.~\cite{kwak2017pedestrian}
explored spatio-temporal feature extraction for real-time prediction.
More recently, Achaji~et~al.~\cite{achaji2022pedestrian} and Lorenzo~et~al.~\cite{lorenzo2021intformer}
applied Transformer attention to capture long-range pedestrian-vehicle interactions,
and Saleh~et~al.~\cite{saleh2019realtime} developed a DenseNet-based
architecture suitable for embedded deployment.
Most closely related to the present study, Sakhai~et~al.~\cite{sakhai2024electronics}
applied spiking neural networks to pedestrian street-crossing detection from
dynamic-vision-sensor data under simulated adverse weather, and released the
accompanying \dvspedx{} synthetic-and-real event dataset~\cite{dvspedx2026}.
We build directly on this line of work, adopting the \dvspedx{} dataset and
treating these event-based results as our primary point of comparison.

\paragraph{Attention and graph-based methods.}
Kotseruba~et~al.~\cite{kotseruba2021benchmark} conducted a thorough evaluation of
intent-prediction backbones on JAAD, finding that CNN-LSTM variants with
attention achieve 78--85\% accuracy.  Graph Convolutional Networks (GCNs)~\cite{kipf2017gcn}
and Spatial-Temporal GCNs~\cite{yan2018stgcn} have been applied to skeletal
action representations; pedestrian-vehicle interaction
graphs~\cite{cadena2019pedestrian} improve robustness to partial occlusion.
\textit{PedFormer}~\cite{pedformer2023} adapts the Transformer
architecture~\cite{vaswani2017attention,bahdanau2015attention} to pedestrian trajectory and intent
prediction, reporting state-of-the-art results in the 85--90\% range on JAAD
under standard protocols.  Vision Transformers~\cite{dosovitskiy2021vit} applied
to pedestrian bounding box crops represent another direction enabled by
large-scale pre-training.

\paragraph{Multi-modal approaches.}
Several works fuse RGB appearance with auxiliary signals such as optical
flow~\cite{dosovitskiy2015flownet,simonyan2014twostream},
skeleton keypoints~\cite{yang2021pedestrian,cao2017openpose}, or vehicle ego-motion.
Video-based 3D convolutional networks~\cite{tran2015c3d} capture short-range
spatio-temporal patterns but require dense RGB input.
Object detection backbones~\cite{ren2015faster,redmon2016yolo} provide bounding
box localisation as input features.
Fusion consistently outperforms single-modality systems but increases compute and
sensor requirements.  Critically, \emph{all} of these approaches depend on
frame cameras; none have been demonstrated with event-based sensors.
Muhammad~et~al.~\cite{muhammad2020deep} provide a comprehensive survey of deep
learning methods for autonomous driving, highlighting the gap in event-based
pedestrian intent research.

\subsection{Event-Based Datasets and Simulators}

\paragraph{\dvspedx{}.}
The \dvspedx{} dataset~\cite{dvspedx2026} provides simulated DVS sequences from
a CARLA-based autonomous driving simulator with labelled pedestrian actions.
It includes both normal-weather and adverse-weather (rain, fog, night-time)
splits, making it uniquely suited to studying robustness.  We use \dvspedx{} as
a complementary training and evaluation source alongside JAAD.

\paragraph{v2e simulator.}
Hu~et~al.~\cite{hu2021v2e} proposed \vte{}, a software pipeline that
generates synthetic DVS events from standard video by modelling the
log-domain photoreceptor response, threshold mismatch, and bandwidth
limiting.  Unlike purely kinematic simulators, \vte{} accounts for photon
noise and sensor non-idealities.  It has been used to extend frame-based
datasets to the DVS domain without physical sensor acquisition.

\paragraph{Other event datasets.}
N-MNIST~\cite{orchard2015nmnist} and N-Caltech101 applied saccadic camera
motion to standard image databases.  DDD17~\cite{binas2017ddd17} provides
real DVS recordings from a vehicle but lacks pedestrian intent labels.
DSEC~\cite{gehrig2021dsec} focuses on optical flow estimation.
HATS~\cite{sironi2018hats} introduced a histogram-based event surface representation
for object classification, while Cannici~et~al.~\cite{cannici2019attention} proposed
attention-based asynchronous event processing.
Messikommer~et~al.~\cite{messikommer2020eventasync} studied asynchronous graph-based
event representations, and the N-ImageNet benchmark~\cite{kim2021nImageNet} enables
large-scale event-camera recognition evaluation.
Event-based object detection~\cite{perot2020learning} further extends the scope.
The ESIM simulator~\cite{rebecq2018esim} provides an alternative physics-based
event generation approach to \vte{}.  Standard autonomous driving datasets
(KITTI~\cite{geiger2012kitti}, nuScenes~\cite{caesar2020nuscenes},
Cityscapes~\cite{cordts2016cityscapes}, CARLA~\cite{dosovitskiy2017carla})
use frame cameras exclusively.
\emph{None} of the above provide crossing-intent binary labels paired with DVS
event streams from real or realistic pedestrian footage, motivating our
JAAD$+$\vte{} conversion.

\subsection{Spiking Neural Networks for Vision}

\paragraph{ANN-to-SNN conversion.}
Rueckauer~et~al.~\cite{rueckauer2017conversion} showed that deep CNNs can be
converted to SNNs by mapping ReLU activations to rate-coded spikes.  Accuracy
is well-preserved at the cost of longer inference horizons (hundreds of
timesteps), incompatible with the short clips ($T{=}9$ frames) used here.

\paragraph{Direct training with surrogate gradients.}
Neftci~et~al.~\cite{neftci2019surrogate} unified surrogate-gradient methods
for backpropagating through spike functions; Zenke~et~al.~\cite{zenke2018superspike}
introduced the SuperSpike learning rule for single-spike timing.
Wu~et~al.~\cite{wu2018snn_backprop} demonstrated spatio-temporal backpropagation
for SNNs and Lee~et~al.~\cite{lee2016training} showed direct training
with learnable thresholds.
SpikingJelly~\cite{fang2023spikingjelly}
provides a PyTorch~\cite{paszke2019pytorch} framework for this paradigm.  We follow
this approach, implementing the surrogate directly via a custom
\texttt{torch.autograd.Function} (Eq.~\ref{eq:surrogate}) rather than an
external library, enabling full control over the gradient shape.
Hybrid SNN-ANN approaches~\cite{rathi2020hybrid,sengupta2019snncnn} combine
the best of both paradigms, and BatchNorm for SNNs~\cite{kim2020batchnorm_snn}
stabilises training of deep spiking networks.
Bellec~et~al.~\cite{bellec2018long} extended LIF networks with adaptation
currents for long-term temporal memory, and Tavanaei~et~al.~\cite{tavanaei2019deep}
surveyed the landscape of deep learning with SNNs.
Pfeiffer and Pfeil~\cite{pfeiffer2018deeplearningsnn} provided a comprehensive
review of deep SNN methodologies, and Diehl and Cook~\cite{diehl2015unsupervised}
demonstrated unsupervised spike-timing-dependent plasticity (STDP)-based SNN learning.
Tan~et~al.~\cite{tan2021spiking} further explored spiking activation mechanisms
for efficient inference.

\paragraph{Convolutional SNNs on DVS data.}
Zhu~et~al.~\cite{zhu2018evflownet} and Amir~et~al.~\cite{amir2017gesture}
applied convolutional processing to event streams for optical flow and gesture
recognition, respectively.  These works demonstrate the superiority of spatial
convolution over flat encodings for event data, which we confirm in the
context of pedestrian intent.

\subsection{Comparison Summary}

Table~\ref{tab:related} positions our work among representative prior systems.
Our method is the first to combine a DVS event representation derived from
JAAD RGB footage with a convolutional SNN trained end-to-end.

\begin{table*}[t]
\centering
\caption{Comparison with representative prior works on pedestrian intent
  prediction and DVS-based SNN classification.
  \textbf{Bold}: best or joint-best per column where comparable.
  ``---'': not applicable or not reported.
  $\dagger$: results on JAAD under standard frame-based evaluation protocol.
  $\ddagger$: our method, DVS input, clip-level evaluation.
  $\S$: the original work reports AUROC and F-score rather than top-line
  accuracy; we list its best CARLA-DVS AUROC ($\times100$) and F-score.}
\label{tab:related}
\footnotesize
\setlength{\tabcolsep}{3pt}
\renewcommand{\arraystretch}{1.12}
\begin{tabularx}{\textwidth}{@{}>{\raggedright\arraybackslash}p{3.05cm} >{\raggedright\arraybackslash}p{2.15cm} >{\raggedright\arraybackslash}p{2.40cm} >{\raggedright\arraybackslash}p{2.20cm} >{\centering\arraybackslash}p{1.95cm} >{\centering\arraybackslash}p{1.15cm} >{\centering\arraybackslash}p{0.70cm} >{\centering\arraybackslash}p{0.70cm} >{\centering\arraybackslash}p{1.25cm}@{}}
\toprule
\textbf{Method} & \textbf{Input} & \textbf{Network} & \textbf{Dataset}
  & \makecell{\textbf{Acc.}\\\textbf{(\%)}} & \textbf{F1}
  & \textbf{DVS} & \textbf{SNN} & \makecell{\textbf{Multi-}\\\textbf{Domain}} \\
\midrule
Rasouli~et~al.~\cite{rasouli2017anticipating} (2017)
  & RGB & CNN+LSTM & JAAD$^\dagger$ & ${\sim}76$ & --- & \xmark & \xmark & \xmark \\
Kotseruba~et~al.~\cite{kotseruba2021benchmark} (2021)
  & RGB+Pose & CNN+LSTM+Attn & JAAD$^\dagger$ & ${\sim}83$ & --- & \xmark & \xmark & \xmark \\
Yang~et~al.~\cite{yang2021pedestrian} (2021)
  & RGB+Skeleton & Graph CNN & JAAD$^\dagger$ & ${\sim}85$ & --- & \xmark & \xmark & \xmark \\
PedFormer~\cite{pedformer2023} (2023)
  & RGB & Transformer & JAAD$^\dagger$ & ${\sim}88$ & --- & \xmark & \xmark & \xmark \\
\midrule
Amir~et~al.~\cite{amir2017gesture} (2017)
  & DVS & SNN (IBM TrueNorth) & DVS Gestures & 96.5 & --- & \cmark & \cmark & \xmark \\
Sakhai~et~al.~\cite{sakhai2024electronics} (2024)
  & DVS (CARLA+JAAD) & SNN (TEBN) & \dvspedx{} & $95.4^{\S}$ & $0.755^{\S}$ & \cmark & \cmark & \cmark \\
\dvspedx{}~\cite{dvspedx2026} (2026)
  & DVS (CARLA+JAAD) & CNN/SNN baselines & \dvspedx{} & ${\sim}90$ & --- & \cmark & \cmark & \cmark \\
\midrule
\textbf{Ours}~(Conv-SNN+Aug)
  & DVS (JAAD+CARLA) & \textbf{Conv-SNN} & JAAD+\dvspedx{}$^\ddagger$
  & \textbf{95.83} (JAAD) / \textbf{97.79} (CARLA) & \textbf{0.9695} / \textbf{0.9478}
  & \cmark & \cmark & \cmark \\
\bottomrule
\end{tabularx}
\normalsize
\end{table*}

\section{DVS Generation Pipeline}
\label{sec:dvs}

\subsection{JAAD Source Data}

The JAAD dataset~\cite{rasouli2019jaad} comprises 346 short video clips of
pedestrians filmed from an ego-vehicle camera in European and North American
urban environments.  Clips are annotated at the pedestrian level with a binary
label: \emph{crossing} (the pedestrian crosses or begins to cross the lane of
the ego-vehicle) or \emph{non-crossing} (the pedestrian stops, waits, or
moves parallel to the vehicle).  Labels are stable across all frames of a
clip: a clip is crossing if the pedestrian commits to crossing during that
clip.

\subsection{\dvspedx{} Dataset}
\label{sec:dvspedx}
\begin{figure*}[tp]
  \centering
  \safeincludegraphics[width=0.98\textwidth]{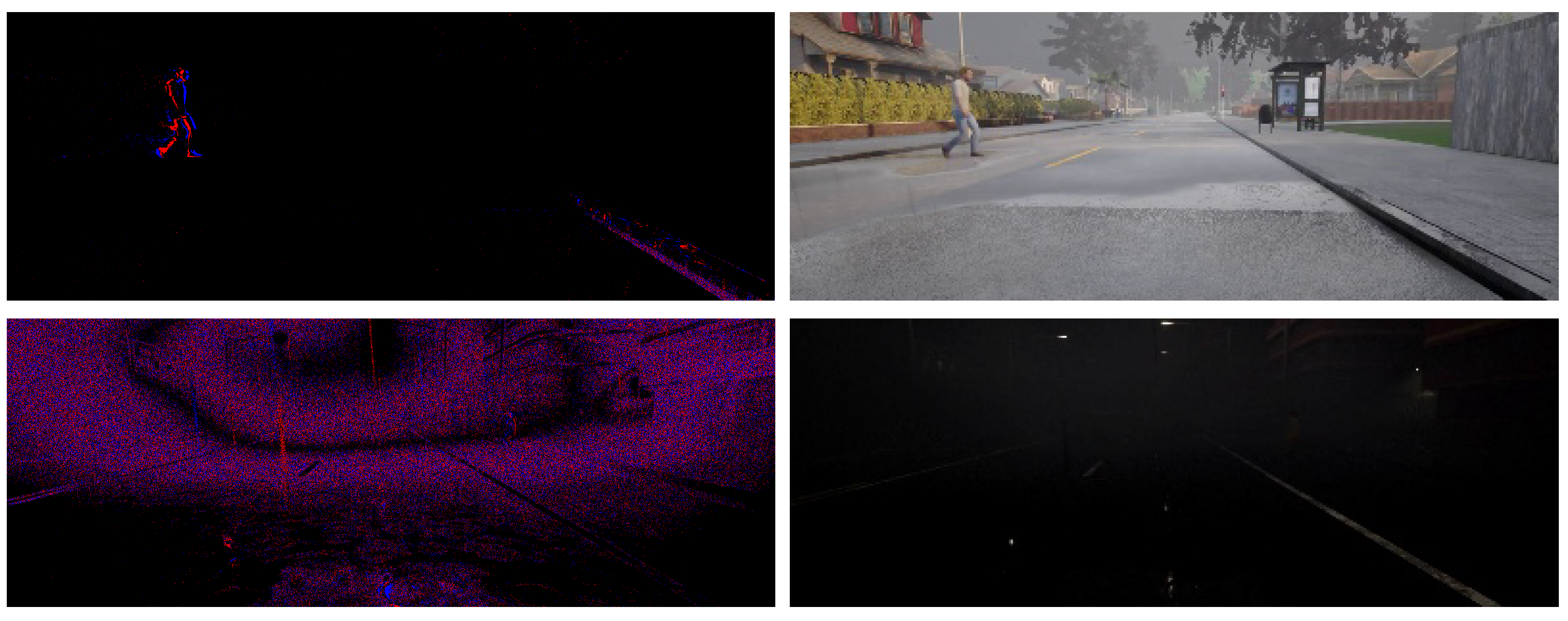}
  \caption{Example frames illustrating the impact of adverse weather on RGB and
    the complementary robustness of DVS. Reproduced from~\cite{sakhai2024electronics}
    (MDPI Electronics 2024, 13(21), 4280).}
  \label{fig:mdpi_dvs_rgb_weather}
\end{figure*}

\begin{figure*}[tp]
  \centering
  \safeincludegraphics[width=0.92\textwidth]{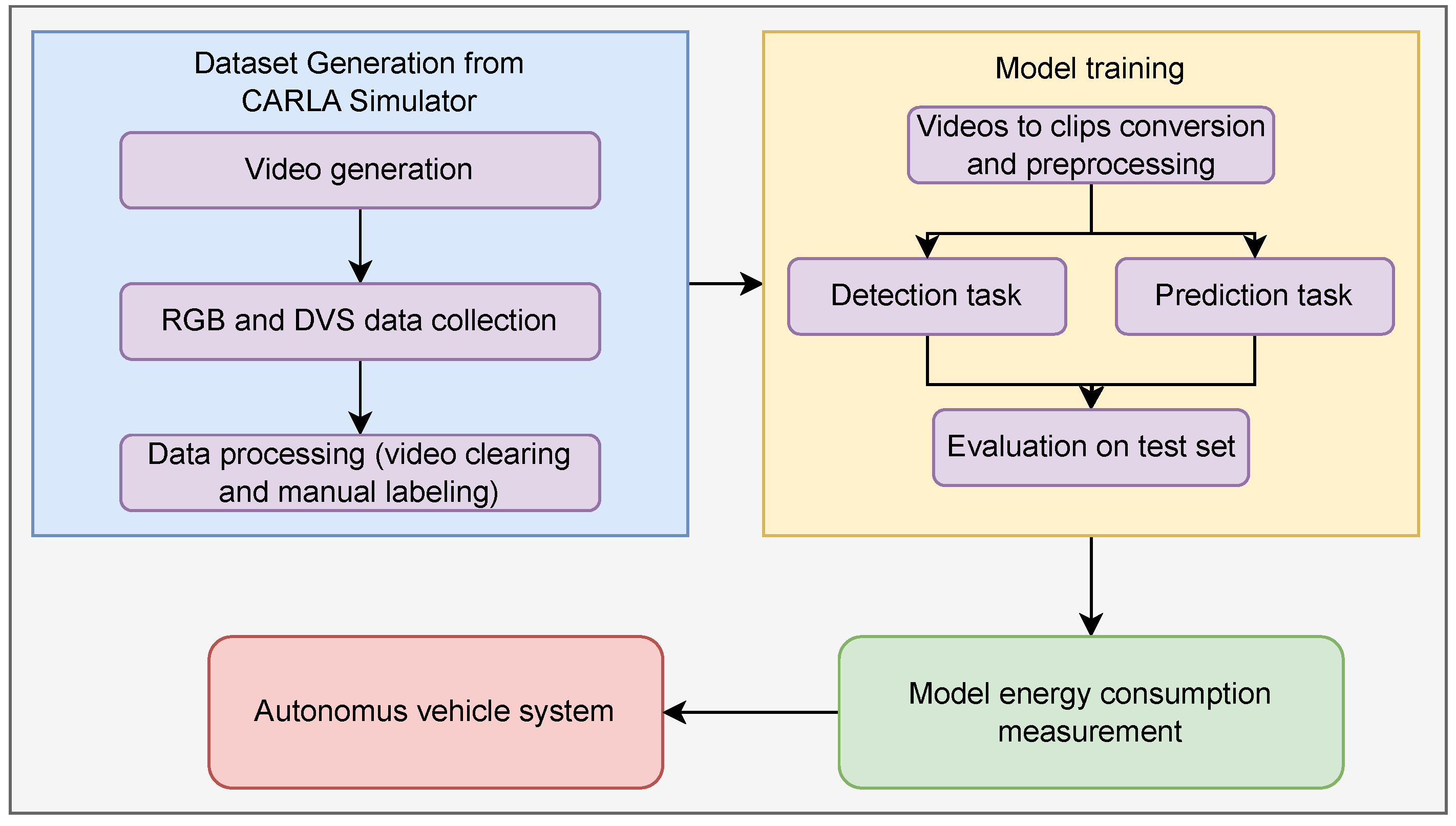}
  \caption{Reference pipeline (simulation $\rightarrow$ preprocessing
    $\rightarrow$ model training/evaluation) for DVS-based pedestrian analysis.
    We adapt this structure to our \vte{}-based JAAD conversion, CARLA/\dvspedx{}
    multi-domain training, and Conv-SNN classification pipeline.
    Reproduced from~\cite{sakhai2024electronics}.}
  \label{fig:mdpi_pipeline}
\end{figure*}

\begin{figure*}[tp]
  \centering
  \begin{subfigure}[t]{0.82\textwidth}
    \centering
    \safeincludegraphics[width=\textwidth]{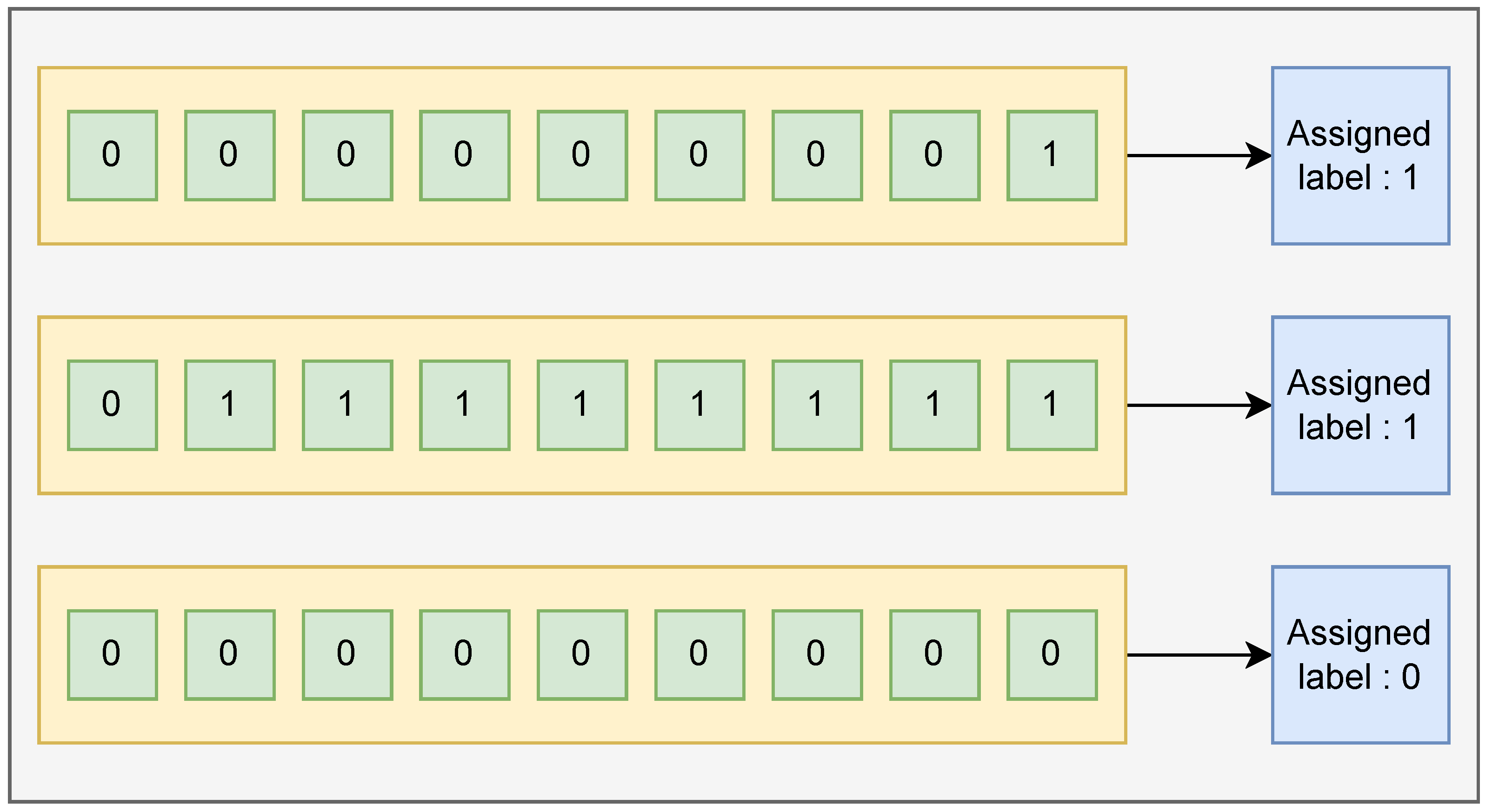}
    \caption{Clip labelling example (detection-style).}
    \label{fig:mdpi_label_detect}
  \end{subfigure}

  \vspace{8pt}

  \begin{subfigure}[t]{0.82\textwidth}
    \centering
    \safeincludegraphics[width=\textwidth]{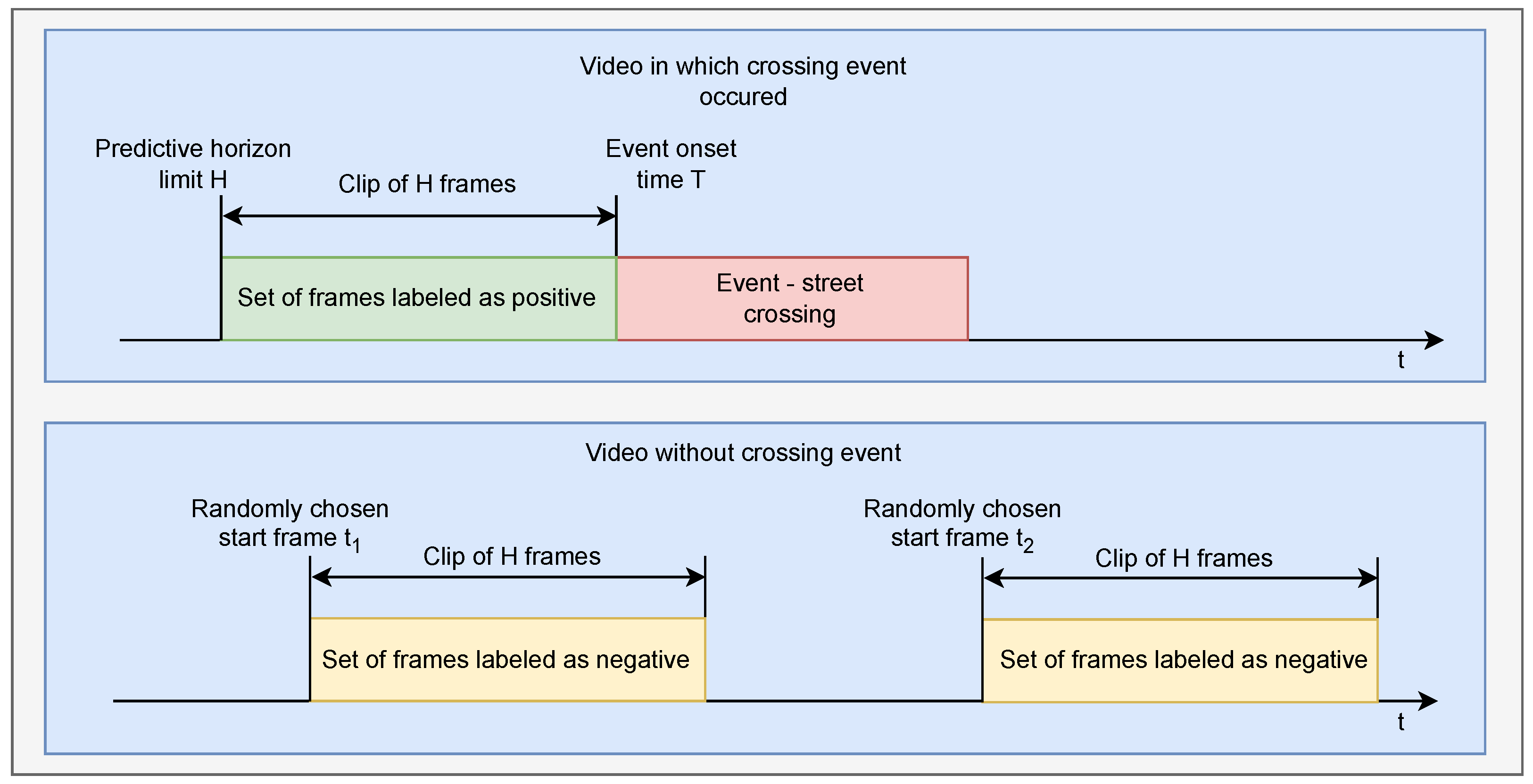}
    \caption{Frame labelling depiction (prediction-style).}
    \label{fig:mdpi_label_predict}
  \end{subfigure}
  \caption{Label assignment illustrations from prior DVS+SNN pedestrian work.
    We follow an explicit clip construction and labelling protocol in our intent
    classification setting (crossing vs.\ non-crossing), with clip-consistent
    augmentations applied across $T$ timesteps. Reproduced from~\cite{sakhai2024electronics}.}
  \label{fig:mdpi_labeling}
\end{figure*}

We obtain the synthetic and simulated portion of our multi-domain training data
from \dvspedx{}~\cite{dvspedx2026}, a dataset designed for pedestrian
street-crossing analysis under both normal and adverse weather. \dvspedx{}
provides event-based sequences generated in the CARLA simulator with paired RGB
frames, temporally accumulated event frames, and frame-level binary labels
(crossing vs.\ non-crossing). In addition, \dvspedx{} includes a real-world
component derived from the \jaad{} dashcam dataset and converted to events via
\vte{}, enabling controlled evaluation of sim-to-real transfer. We follow the
dataset's official splits where applicable, and treat JAAD-derived event clips
and CARLA-derived event sequences as complementary domains during training and
testing. Figure~\ref{fig:mdpi_dvs_rgb_weather} illustrates the degradation of
RGB imagery under adverse weather and the complementary robustness of DVS
sensing that motivates this multi-domain design.
Figure~\ref{fig:mdpi_pipeline} shows the reference
simulation-to-evaluation pipeline that our approach adapts.
Figure~\ref{fig:mdpi_labeling} depicts the clip- and frame-level label
assignment protocols from prior DVS+SNN pedestrian work that inform our
labelling scheme.

Each clip is stored as a sequence of RGB PNG frames.  The clip length varies
from tens to several hundred frames depending on pedestrian behaviour and
scene complexity.

\subsection{v2e DVS Simulation}
\label{sec:v2e_detail}

We convert each JAAD RGB clip to a synthetic DVS event stream using the
\vte{} simulator~\cite{hu2021v2e}, specifically its \texttt{EventEmulator} and
\texttt{EventRenderer} pipeline run in headless mode (all GUI dependencies
mocked).  The conversion proceeds frame-by-frame without any intermediate
video encoding.

\paragraph{Photoreceptor model.}
The DVS pixel response is modelled as a first-order low-pass filter in the
log-luminance domain.  For a greyscale pixel value $I(t)$, define
\begin{equation}
  L(t) = \ln\!\bigl(I(t) + \epsilon\bigr), \quad \epsilon = 10^{-3},
  \label{eq:log_lum}
\end{equation}
and let $\hat{L}(t)$ be the low-pass filtered version with cutoff
$f_c = 15$\,Hz (first-order IIR):
\begin{equation}
  \hat{L}(t) = \alpha\,\hat{L}(t{-}1) + (1-\alpha)\,L(t),
  \quad \alpha = e^{-2\pi f_c \Delta t}.
  \label{eq:lpf}
\end{equation}

\paragraph{Event generation.}
An ON event fires when the change in filtered log-luminance exceeds the
positive threshold and an OFF event when it falls below the negative threshold:
\begin{align}
  \text{ON:} &\quad \hat{L}(t) - \hat{L}(t_{\text{ref}}) >
    \theta_+ + \mathcal{N}(0,\sigma^2), \label{eq:on} \\
  \text{OFF:} &\quad \hat{L}(t) - \hat{L}(t_{\text{ref}}) <
    -\theta_- - \mathcal{N}(0,\sigma^2), \label{eq:off}
\end{align}
where $t_{\text{ref}}$ is the timestamp of the last event at that pixel.
We use the calibrated parameters from the \vte{} parameter reference:
\begin{center}
\begin{tabular}{ll}
  $\theta_+ = \theta_- = 0.15$ & (log-intensity units) \\
  $\sigma = 0.03$ & (threshold noise std.) \\
  $f_c = 15$\,Hz & (photoreceptor cutoff) \\
  $\lambda_{\text{leak}} = 0.01$\,Hz & (leak rate) \\
  $\lambda_{\text{shot}} = 0.0$\,Hz & (shot noise, disabled)
\end{tabular}
\end{center}

\paragraph{Frame rendering.}
Events within one inter-frame interval are accumulated and rendered to a
greyscale image using \texttt{ExposureMode.SOURCE}: pixels with ON events
are set to 255, OFF events to 0, and inactive pixels to 128 (neutral grey).
The output resolution is set to half the input resolution (width~$/2\,\times$~height~$/2$)
to reduce redundancy.  All DVS frames are then resized to
$64 \times 64$ pixels during dataset loading.

\paragraph{Empty-frame filtering.}
A frame is considered \emph{empty} (no meaningful events) if:
\begin{equation}
  \sigma_{\text{img}} \leq 8 \quad \text{and} \quad
  |\mu_{\text{img}} - 128| \leq 12,
  \label{eq:empty}
\end{equation}
where $\mu_{\text{img}}$ and $\sigma_{\text{img}}$ are the per-frame pixel mean
and standard deviation.  Empty frames are discarded before training.  This
prevents the network from learning a trivial shortcut: static scenes produce
no events, but label-derived shortcuts could allow the model to classify by
clip identity rather than motion content.

\subsection{CARLA \dvspedx{} Source}

The \dvspedx{} dataset~\cite{dvspedx2026} provides simulated DVS frames from
CARLA, a physically based urban driving simulator.  We use two splits:

\begin{itemize}[leftmargin=1.4em,topsep=2pt,itemsep=2pt]
  \item \textbf{Normal}: clear weather, daytime scenes.
  \item \textbf{Adverse weather}: rain, fog, and night-time conditions.
\end{itemize}

Frames are organised in sequence directories; each filename encodes the
frame index and binary crossing label ($N$-$Y$.png, where $Y \in \{0,1\}$).
Clip labels are defined as the maximum label within the clip window (crossing
if any frame is labelled crossing), which correctly handles clips that span
the onset of pedestrian movement.  Frames are pre-resized to $64 \times 64$
in a cache to avoid loading large originals at training time.

\begin{figure}[tbp]
\centering
\resizebox{\columnwidth}{!}{%
\begin{tikzpicture}[
  font=\footnotesize\sffamily,
  node distance=7mm and 11mm,
  box/.style={draw=black!55, line width=0.5pt, rounded corners=2pt,
              text centered, align=center,
              minimum height=9mm, inner sep=3pt, minimum width=27mm},
  input/.style={box, fill=blue!9},
  proc/.style={box, fill=cyan!13},
  merge/.style={box, fill=green!16, draw=green!50!black},
  model/.style={box, fill=orange!20, draw=orange!60!black},
  outbox/.style={box, fill=red!15, draw=red!60!black},
  grp/.style={draw=black!30, dash pattern=on 2pt off 1.5pt,
              rounded corners=3pt, inner sep=3.5mm},
  ar/.style={-{Stealth[length=2.4mm,width=1.9mm]},
             line width=0.8pt, draw=black!68, rounded corners=2pt},
]
\node[input] (jaad) {JAAD RGB clips\\($346$)};
\node[proc,  below=of jaad] (v2e)  {\vte{} emulator\\$\theta{=}0.15,\ f_c{=}15$\,Hz};
\node[input, below=of v2e]  (jdvs) {Synthetic DVS\\$64{\times}64$};
\node[input, right=of jaad] (carla) {CARLA \dvspedx{}\\normal $+$ adverse};
\node[input, below=of carla] (cache) {Pre-cached frames\\$64{\times}64$};
\node[merge, below=15mm of jdvs, xshift=19mm] (merge)
  {Combined dataset\\JAAD ($6{\times}$) $+$ CARLA};
\node[model, below=of merge] (model) {Conv-SNN $+$ aug\\$T{=}9,\ B{=}32$};
\node[outbox, below=of model] (out)   {Crossing /\\non-crossing};
\draw[ar] (jaad) -- (v2e);
\draw[ar] (v2e)  -- (jdvs);
\draw[ar] (carla) -- (cache);
\draw[ar] (jdvs.south)  -- ++(0,-4mm) -| (merge.north west);
\draw[ar] (cache.south) -- ++(0,-4mm) -| (merge.north east);
\draw[ar] (merge) -- (model);
\draw[ar] (model) -- (out);
\begin{scope}[on background layer]
  \node[grp, fill=blue!3, fit=(jaad)(v2e)(jdvs)] {};
  \node[grp, fill=cyan!4, fit=(carla)(cache)] {};
\end{scope}
\end{tikzpicture}%
}
\caption{End-to-end pipeline.  JAAD RGB clips are converted to synthetic
  DVS via \vte{}; CARLA \dvspedx{} sequences are loaded from a pre-cached
  directory.  Both sources feed Conv-SNN after $6\times$ JAAD oversampling.}
\label{fig:pipeline}
\end{figure}
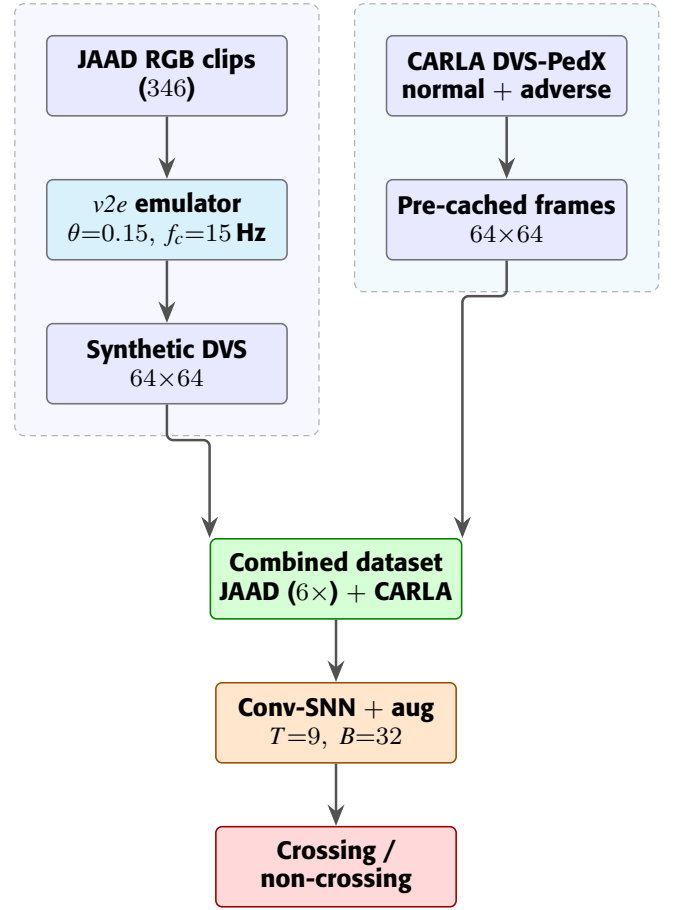

Figure~\ref{fig:pipeline} summarises the full data pipeline.

\section{Proposed Method}
\label{sec:method}

\subsection{Problem Formulation}

Given a clip of $T$ consecutive DVS frames
$\mathbf{X} = \{X_1, X_2, \ldots, X_T\}$ where $X_t \in \mathbb{R}^{H \times W}$,
we seek a classifier $f_\theta: \mathbb{R}^{T \times H \times W} \to \{0, 1\}$
that predicts $y \in \{0, 1\}$ where $y=1$ denotes \emph{crossing} and $y=0$
denotes \emph{non-crossing}.  We parameterise $f_\theta$ as a convolutional
SNN and learn $\theta$ by minimising a class-weighted cross-entropy loss.

\subsection{Leaky-Integrate-and-Fire Neuron}
\label{sec:lif}

The fundamental computational unit is the Leaky-Integrate-and-Fire (LIF)
neuron.  In continuous time, the membrane potential $\vmem(t)$ evolves as:
\begin{equation}
  \vtau \frac{\mathrm{d}\vmem}{\mathrm{d}t}
    = -\!\left(\vmem - \vleak\right) + I_{\mathrm{syn}}(t),
  \label{eq:lif_cont}
\end{equation}
where $\vtau$ is the membrane time constant, $\vleak$ is the resting
potential, and $I_{\mathrm{syn}}$ is the synaptic input current.

\paragraph{Discrete-time update.}
In our implementation, a single Euler step with step size $\Delta t = 1$ frame
gives:
\begin{equation}
  \begin{aligned}
    \Delta\vmem &= \frac{-(\vmem - \vleak) + I_{\mathrm{syn}}}{\vtau}, \\
    \vmem &\leftarrow \vmem + \Delta\vmem.
  \end{aligned}
  \label{eq:lif_update}
\end{equation}
A spike fires whenever $\vmem \geq \vth$:
\begin{equation}
  s = \mathcal{H}(\vmem - \vth),
  \quad
  \mathcal{H}(x) =
  \begin{cases}
    1 & x \geq 0 \\
    0 & x < 0
  \end{cases}
  \label{eq:spike_fn}
\end{equation}
After firing, the membrane potential undergoes a \emph{soft reset}:
\begin{equation}
  \vmem \leftarrow \vmem\cdot(1 - s) + \vleak\cdot s,
  \label{eq:reset}
\end{equation}
which sets $\vmem = \vleak$ for fired neurons while leaving non-fired neurons
unchanged.  Compared to hard reset ($\vmem \leftarrow \vleak$), soft reset
preserves the sub-threshold dynamics of neurons that fire just over threshold,
stabilising gradient flow.

The LIF parameters used throughout are:
\begin{equation}
  \vth = 0.5, \quad \vleak = 0.0, \quad \vtau = 2.0.
  \label{eq:lif_params}
\end{equation}
The threshold $\vth = 0.5$ was chosen so that normalised inputs
($\sim\mathcal{N}(0,1)$ after BatchNorm) yield a spike rate of
approximately 15--20\%, consistent with biologically plausible sparse coding.

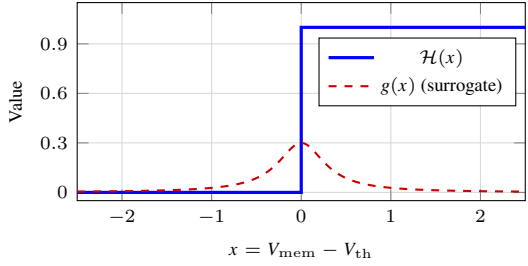
\begin{figure}[tbp]
\centering
\begin{tikzpicture}
\begin{axis}[
  width=0.88\linewidth, height=4.2cm,
  xlabel={$x = \vmem - \vth$},
  ylabel={Value},
  xmin=-2.5, xmax=2.5,
  ymin=-0.05, ymax=1.15,
  xtick={-2,-1,0,1,2},
  ytick={0,0.3,0.6,0.9},
  grid=both, grid style={gray!30},
  legend style={at={(0.98,0.65)},anchor=east,font=\scriptsize},
  tick label style={font=\scriptsize},
  label style={font=\scriptsize},
]
\addplot[blue, very thick, const plot, domain=-2.5:2.5]
  coordinates {(-2.5,0)(0,0)(0,1)(2.5,1)};
\addlegendentry{$\mathcal{H}(x)$}
\addplot[red!80!black, thick, dashed, domain=-2.5:2.5, samples=200]
  {0.3/(1+(3.14159*x)^2)};
\addlegendentry{$g(x)$ (surrogate)}
\end{axis}
\end{tikzpicture}
\caption{Spike function $\mathcal{H}(x)$ (blue) and its surrogate
  gradient $g(x)$ (red dashed) used for backpropagation.  The surrogate
  peaks at $x=0$ with value $0.3$ and decays symmetrically.}
\label{fig:surrogate}
\end{figure}

\paragraph{Surrogate gradient.}
Because $\mathcal{H}$ has zero gradient almost everywhere, direct
backpropagation through $s$ is impossible.  We adopt the standard
arctangent-derived surrogate~\cite{neftci2019surrogate}; we do not claim it
as a contribution, but two implementation choices are specific to our setting:
\begin{equation}
  \frac{\partial \mathcal{H}(x)}{\partial x}
  \approx g(x) = \frac{0.3}{1 + \left(\pi x\right)^2}.
  \label{eq:surrogate}
\end{equation}
First, rather than relying on an external SNN library, the surrogate is
implemented directly as a custom \texttt{torch.autograd.Function}: the forward
pass emits the hard spike $\mathcal{H}(x)$ while the backward pass substitutes
$g(x)$.  This gives full control over the gradient shape and removes any
framework dependency in the spike path.  Second, the peak ($0.3$) and width
(the $\pi$ scaling) are tuned to our threshold/normalisation regime
($\vth=0.5$ on BatchNorm-standardised inputs, Eq.~\ref{eq:lif_params}).  The
strongest gradient is thus delivered to neurons firing near threshold, while
clearly sub- or super-threshold neurons receive vanishing gradient.  Figure~\ref{fig:surrogate}
shows $g(x)$ overlaid with $\mathcal{H}(x)$: it is smooth, symmetric about
$x=0$, and decays to zero for large $|x|$, preventing gradient blow-up.

\subsection{Convolutional LIF Block}
\label{sec:convlif}

The use of convolutional feature extraction is motivated by decades of
evidence that hierarchical spatial filtering outperforms flat representations
for visual tasks~\cite{lecun1998gradient,krizhevsky2012alexnet,simonyan2015vgg,he2016resnet,szegedy2015inception}.
Network-in-network pooling~\cite{lin2014network} further consolidates spatial
features before fully connected classification.

A \textbf{ConvLIF} block processes a feature map
$\mathbf{s}^{(t)} \in \mathbb{R}^{C_{\mathrm{in}} \times H \times W}$
at timestep $t$ as follows:
\begin{align}
  \mathbf{z}^{(t)} &= \mathrm{MaxPool}\!\left(
    \mathrm{BN}\!\left(\mathrm{Conv}_{3\times3}\!\left(\mathbf{s}^{(t)}\right)
    \right)\right), \label{eq:convlif_pre} \\
  \mathbf{s}^{(t+1)}, \mathbf{v}^{(t+1)} &=
    \mathrm{LIF}\!\left(\mathbf{z}^{(t)}, \mathbf{v}^{(t)}\right),
  \label{eq:convlif_lif}
\end{align}
where BatchNorm~\cite{ioffe2015batchnorm} (BN) is applied across the batch dimension at each timestep,
$\mathrm{MaxPool}(2\times2)$ reduces spatial resolution by $2\times$, and LIF
integrates the pre-activation $\mathbf{z}$ into the membrane potential
$\mathbf{v}$.  Convolutional weights use Kaiming uniform
initialisation~\cite{he2015kaiming} appropriate for non-linearity ``relu''
(which approximates the spike function in expectation); biases use Xavier
uniform initialisation~\cite{glorot2010xavier}.

\subsection{Full Conv-SNN Architecture}
\label{sec:arch}

The complete \csnn{} processes a clip
$\mathbf{X} \in \mathbb{R}^{T \times H \times W}$ frame-by-frame.
All membrane potentials are initialised to zero at the start of each clip.

\paragraph{Architecture block diagram.}
Table~\ref{tab:arch_params} illustrates the full architecture:
Table~\ref{tab:arch} lists the layer-by-layer operations, and
Table~\ref{tab:params} the per-layer parameter counts.  For each timestep
$t = 1, \ldots, T$:

\begin{enumerate}[leftmargin=1.4em,topsep=2pt,itemsep=2pt]
  \item Input frame $X_t \in \mathbb{R}^{1 \times 64 \times 64}$.
  \item \textbf{ConvLIF-1}: Conv$(1{\to}16, 3{\times}3)$ + BN + MaxPool$(2{\times}2)$
    + LIF $\to$ spike map $\mathbb{R}^{16 \times 32 \times 32}$.
  \item \textbf{ConvLIF-2}: Conv$(16{\to}32, 3{\times}3)$ + BN + MaxPool$(2{\times}2)$
    + LIF $\to$ $\mathbb{R}^{32 \times 16 \times 16}$.
  \item \textbf{ConvLIF-3}: Conv$(32{\to}64, 3{\times}3)$ + BN + MaxPool$(2{\times}2)$
    + LIF $\to$ $\mathbb{R}^{64 \times 8 \times 8}$.
  \item \textbf{Flatten}: $\mathbb{R}^{64 \times 8 \times 8} \to \mathbb{R}^{4096}$.
  \item \textbf{FC-LIF}: Linear$(4096{\to}256)$ + BN1d + LIF
    $\to$ spike vector $\mathbb{R}^{256}$.
  \item \textbf{Output}: Linear$(256{\to}2)$; logits accumulated.
\end{enumerate}

The output logits across all $T$ timesteps are averaged before softmax:
\begin{equation}
  \hat{\mathbf{y}} = \frac{1}{T}\sum_{t=1}^{T}
    \mathbf{W}_o\,\mathbf{s}_{\mathrm{FC}}^{(t)},
  \label{eq:accum}
\end{equation}
where $\mathbf{W}_o \in \mathbb{R}^{2 \times 256}$ and
$\mathbf{s}_{\mathrm{FC}}^{(t)}$ is the FC-LIF spike at time $t$.

\begin{table*}[t]
\centering
\setlength{\tabcolsep}{4pt}
\renewcommand{\arraystretch}{1.18}
\begin{subtable}[t]{0.56\textwidth}
\centering
\begin{tabular}{clc}
\toprule
\textbf{Block} & \textbf{Operation} & \textbf{Output shape} \\
\midrule
Input      & DVS frame $X_t$                    & $1\times64\times64$ \\
\midrule
ConvLIF-1  & Conv $1{\to}16$, $3\times3$, BN    & $16\times64\times64$ \\
           & MaxPool $2\times2$ + LIF           & $16\times32\times32$ \\
\midrule
ConvLIF-2  & Conv $16{\to}32$, $3\times3$, BN   & $32\times32\times32$ \\
           & MaxPool $2\times2$ + LIF           & $32\times16\times16$ \\
\midrule
ConvLIF-3  & Conv $32{\to}64$, $3\times3$, BN   & $64\times16\times16$ \\
           & MaxPool $2\times2$ + LIF           & $64\times8\times8$   \\
\midrule
Flatten    & Reshape                            & $4096$               \\
FC-LIF     & Linear $4096{\to}256$ + BN1d + LIF & $256$               \\
Output     & Linear $256{\to}2$, avg over $T$   & $2$ (logits)         \\
\bottomrule
\end{tabular}
\caption{Conv-SNN layer-by-layer architecture. Operations in each ConvLIF block
are applied per timestep $t=1,\ldots,T$; the membrane potential $\mathbf{v}$
persists across timesteps.}
\label{tab:arch}
\end{subtable}\hfill
\begin{subtable}[t]{0.42\textwidth}
\centering
\begin{tabular}{lrr}
\toprule
\textbf{Layer} & \textbf{Shape} & \textbf{Params} \\
\midrule
ConvLIF-1 (Conv) & $16{\times}1{\times}3{\times}3$ & 144 \\
ConvLIF-1 (BN)   & $2{\times}16$   & 32 \\
ConvLIF-2 (Conv) & $32{\times}16{\times}3{\times}3$ & 4,608 \\
ConvLIF-2 (BN)   & $2{\times}32$   & 64 \\
ConvLIF-3 (Conv) & $64{\times}32{\times}3{\times}3$ & 18,432 \\
ConvLIF-3 (BN)   & $2{\times}64$   & 128 \\
FC-1 (Linear)    & $256{\times}4096 + 256$ & 1,048,832 \\
BN-FC (BN1d)     & $2{\times}256$  & 512 \\
FC-Out (Linear)  & $2{\times}256 + 2$ & 514 \\
\midrule
\textbf{Total}   & & \textbf{1,073,266} \\
\bottomrule
\end{tabular}
\caption{Layer-wise trainable parameter counts.}
\label{tab:params}
\end{subtable}
\caption{Conv-SNN architecture and parameterisation summary.}
\label{tab:arch_params}
\end{table*}

\paragraph{Parameter count.}
The total number of trainable parameters is $\mathbf{1{,}073{,}266}$ (Table~\ref{tab:params}).

\subsection{Clip-Consistent DVS Augmentation}
\label{sec:aug}

Spatial augmentation of event data requires care: augmentations applied
independently per frame would introduce temporal incoherence, corrupting the
spike dynamics that LIF neurons rely on.  We use three augmentations with
a single random draw per clip, applied \emph{identically} to all $T$ frames:

\begin{enumerate}[leftmargin=1.4em,topsep=2pt,itemsep=1pt]
  \item \textbf{Horizontal flip} (probability $p = 0.5$): DVS ON/OFF polarity
    patterns are laterally symmetric; pedestrians walking left/right are
    equivalent for intent classification.
  \item \textbf{Brightness jitter}: pixel values are multiplied by a factor
    $b \sim \mathcal{U}(0.8, 1.2)$.  This simulates threshold variation
    across different DVS sensor calibrations.
  \item \textbf{Random spatial crop}: a $56{\times}56$ region is cropped
    from a uniformly sampled corner $(i,j)$ where $i,j \sim \mathcal{U}(0,8)$
    (the $64{-}56=8$ pixel margin), then resized back to $64{\times}64$ via
    nearest-neighbour interpolation.  This introduces modest positional
    invariance without distorting the event polarity encoding.
\end{enumerate}

These augmentations follow best practices for visual
recognition~\cite{shorten2019survey,cubuk2019autoaugment}, adapted to the
event-camera domain.  They are applied \emph{only to JAAD clips} during training;
CARLA clips are used without augmentation to preserve the simulator's spatial
statistics.

\subsection{Dataset Construction and Class Balancing}
\label{sec:data}

Three datasets are loaded independently: JAAD DVS (augmented), CARLA DVS
normal, and CARLA DVS adverse weather.  Clips are defined using overlapping
windows of length $T = 9$ with stride $\lfloor T/2 \rfloor = 4$.  Each
dataset is split independently in a \textbf{70/15/15} ratio (train/val/test)
stratified by label, so the class ratio is preserved in each split.

\paragraph{JAAD oversampling.}
The CARLA splits are substantially larger than JAAD.  To prevent CARLA
statistics from dominating learning, we oversample JAAD by a factor
$k = \max\!\left(1,\, \mathrm{round}(N_\text{CARLA} / N_\text{JAAD})\right)$
where $N_\text{CARLA}$ is the combined CARLA training size and $N_\text{JAAD}$
is the JAAD training size.  In our run: $k = 6$, yielding
$4{,}576 \times 6 = 27{,}456$ JAAD training clips with augmentation.  Each
oversampled repeat of a clip receives an independently sampled augmentation,
so the $6\times$ copies are not identical.

\paragraph{Class weights.}
After oversampling, the combined training set has
$31{,}127$ non-crossing clips and $24{,}583$ crossing clips (total $55{,}710$).
Class imbalance is a pervasive challenge in pedestrian datasets~\cite{he2009imbalanced};
rather than synthetic resampling~\cite{chawla2002smote}, we use
inverse-frequency class weights computed as:
\begin{equation}
  w_c = \frac{N_{\text{total}}}{K \cdot N_c},
  \quad c \in \{0, 1\},\; K = 2,
  \label{eq:class_weight}
\end{equation}
giving $w_0 = 0.8949$ and $w_1 = 1.1331$.  These weights are applied to the
cross-entropy loss at each training step.

\subsection{Loss Function}
\label{sec:loss}

The training objective is class-weighted cross-entropy.  For a batch
$\{(\mathbf{X}_i, y_i)\}_{i=1}^B$ with model output
$\hat{\mathbf{y}}_i = f_\theta(\mathbf{X}_i) \in \mathbb{R}^2$:
\begin{equation}
  \mathcal{L} = -\frac{1}{B}\sum_{i=1}^{B}
    w_{y_i} \ln \frac{e^{\hat{y}_{i,y_i}}}{\sum_{c=0}^{1} e^{\hat{y}_{i,c}}},
  \label{eq:loss}
\end{equation}
where $w_{y_i}$ is the class weight for ground-truth class $y_i$.  During
final evaluation the class weights are removed ($w_c = 1$) to obtain
unbiased test metrics.

\subsection{Optimisation}
\label{sec:optim}

We use \textbf{AdamW}~\cite{loshchilov2019adamw}, a decoupled weight-decay
variant of Adam~\cite{kingma2015adam}, as the base optimiser:
\begin{equation}
  \bm{\theta}_{n+1} = \bm{\theta}_n - \eta\,\hat{\mathbf{m}}_n / (\hat{\mathbf{v}}_n^{1/2} + \varepsilon)
    - \eta\,\lambda\,\bm{\theta}_n,
  \label{eq:adamw}
\end{equation}
where $\hat{\mathbf{m}}_n$, $\hat{\mathbf{v}}_n$ are bias-corrected moment
estimates and the last term is the decoupled weight decay.  All parameters are:

\begin{center}
\begin{tabular}{ll}
  Learning rate $\eta$ & $5 \times 10^{-3}$ \\
  $\beta_1, \beta_2$ & $0.9, 0.999$ \\
  $\varepsilon$ & $10^{-8}$ \\
  Weight decay $\lambda$ & $10^{-4}$ \\
  Gradient clip (max norm) & $1.0$ \\
\end{tabular}
\end{center}

Gradients are clipped to a maximum $\ell_2$ norm of $1.0$~\cite{pascanu2013gradient}
to prevent gradient explosions during the early spike-learning phase.
A \textbf{ReduceLROnPlateau} scheduler~\cite{smith2017cyclical} monitors the
validation accuracy (\texttt{mode='max'}) and halves the learning rate
(\texttt{factor=0.5}) whenever it fails to improve for \texttt{patience=2}
consecutive epochs, i.e., $\eta \leftarrow 0.5\,\eta$.
Early stopping~\cite{prechelt1998early} terminates training if validation accuracy does not improve
by at least $\delta = 0.001$ for 5 consecutive epochs, preventing overfitting
without manual epoch selection.

\section{Training Procedure}
\label{sec:training}

\subsection{Hyperparameter Summary}

All hyperparameters are fixed constants in the source code with no search.
Table~\ref{tab:hparams} provides the complete reference.

\begin{table}[htbp]
\centering
\caption{Complete hyperparameter configuration.}
\label{tab:hparams}
{\small%
\setlength{\tabcolsep}{4pt}
\begin{tabularx}{\columnwidth}{@{}l l >{\raggedright\arraybackslash}X@{}}
\toprule
\textbf{Hyperparameter} & \textbf{Value} & \textbf{Justification} \\
\midrule
Random seed          & 42           & Reproducibility \\
Clip length $T$      & 9 frames     & Captures intent gesture duration \\
Frame resolution     & $64{\times}64$ & Cache matches CARLA \dvspedx{} \\
Batch size $B$       & 32           & GPU memory / gradient variance trade-off \\
Max epochs           & 15           & Observed convergence before 15 \\
Learning rate $\eta$ & $5{\times}10^{-3}$ & Aggressive LR for fast spike convergence \\
Weight decay $\lambda$& $10^{-4}$   & L2 regularisation \\
LR patience          & 2 epochs     & Quick LR halving on plateaus \\
LR factor            & 0.5          & Halving \\
Early stopping patience & 5 epochs & Allow short plateaus \\
Min delta $\delta$   & 0.001        & Meaningful improvement threshold \\
JAAD oversample $k$  & 6            & Computed: $\lceil N_\text{CARLA}/N_\text{JAAD}\rceil$ \\
\midrule
LIF $\vth$           & 0.5          & ${\sim}15\%$ spike rate with BN outputs \\
LIF $\vleak$         & 0.0          & Reset to zero \\
LIF $\vtau$          & 2.0          & Half-life ${\approx}1.4$ steps \\
\midrule
Augment: flip prob.  & 0.5          & Bilateral symmetry \\
Augment: brightness  & $[0.8, 1.2]$ & Sensor calibration variation \\
Augment: crop size   & $56{\times}56 \to 64{\times}64$ & 8-pixel margin \\
\midrule
Device               & CPU          & cuBLAS stability; avoids GEMM errors \\
\bottomrule
\end{tabularx}}
\end{table}

\subsection{Training Algorithm}

Algorithm~\ref{alg:train} gives the complete pseudocode for the training loop.

\begin{algorithm}[t]
\caption{Conv-SNN Training with JAAD Oversampling and Class Weighting}
\label{alg:train}
\begin{algorithmic}[1]
\Require Datasets $\mathcal{D}_J$ (JAAD), $\mathcal{D}_C$ (CARLA),
         $\mathcal{D}_A$ (CARLA-Adv)
\Require Hyperparameters: $T,B,\eta,\lambda,k,w_0,w_1,E_{\max}$
\State Initialise Conv-SNN weights $\theta$ (Kaiming/Xavier)
\State Compute class weights $w_0, w_1$ via Eq.~\ref{eq:class_weight}
\State Split each $\mathcal{D}$ 70/15/15 stratified $\Rightarrow$ train/val/test
\State Build $\mathcal{D}_{\mathrm{train}} =
  [\mathcal{D}_{J,\mathrm{train}}]^k
  \cup \mathcal{D}_{C,\mathrm{train}}
  \cup \mathcal{D}_{A,\mathrm{train}}$ \Comment{$k{=}6$, aug each repeat}
\State $\text{best\_acc} \leftarrow 0$;\;
       $p \leftarrow 0$ \Comment{patience counter}
\For{$e = 1$ \textbf{to} $E_{\max}$}
  \State $\mathcal{L}_{\text{epoch}} \leftarrow 0$
  \For{each mini-batch $(\mathbf{X}_b, \mathbf{y}_b) \in \mathcal{D}_{\mathrm{train}}$}
    \State $\hat{\mathbf{y}}_b \leftarrow \mathrm{ForwardConvSNN}(\mathbf{X}_b, \theta)$
    \Comment{Eq.~\ref{eq:accum}}
    \State $\mathcal{L} \leftarrow \mathrm{WeightedCE}(\hat{\mathbf{y}}_b, \mathbf{y}_b, w_0, w_1)$
    \Comment{Eq.~\ref{eq:loss}}
    \State $\theta \leftarrow \theta - \mathrm{AdamW}(\nabla_\theta \mathcal{L})$
    \Comment{clip grad norm ${\le}1.0$}
  \EndFor
  \State $\mathrm{acc}_{\mathrm{val}} \leftarrow \mathrm{Evaluate}(\theta, \mathcal{D}_{\mathrm{val}})$
  \State $\eta \leftarrow \mathrm{ReduceLROnPlateau}(\eta, \mathrm{acc}_{\mathrm{val}})$
  \If{$\mathrm{acc}_{\mathrm{val}} > \text{best\_acc} + 0.001$}
    \State $\theta^* \leftarrow \theta$;\;
           $\text{best\_acc} \leftarrow \mathrm{acc}_{\mathrm{val}}$;\;
           $p \leftarrow 0$
  \Else
    \State $p \leftarrow p + 1$
    \If{$p \geq 5$} \textbf{break} \EndIf
  \EndIf
\EndFor
\State \Return $\theta^*$ \Comment{best validation checkpoint}
\end{algorithmic}
\end{algorithm}

\section{Experiments}
\label{sec:experiments}

\subsection{Dataset Statistics}

Table~\ref{tab:dataset} summarises the dataset composition.
All clips are formed with a sliding window of length $T{=}9$ and
stride 4 (half the clip length).

\begin{table*}[t]
\centering
\caption{Dataset statistics before splitting and oversampling.}
\label{tab:dataset}
{\small%
\setlength{\tabcolsep}{6pt}
\begin{tabular}{@{}lcccc@{}}
\toprule
\textbf{Source} & \textbf{Type} & \textbf{Classes} & \textbf{Clips} & \textbf{Resolution} \\
\midrule
JAAD DVS & Synthetic (\vte{}) & Crossing / Non-crossing & 4,576+ & $64{\times}64$ \\
CARLA DVS (normal)  & Simulated & Crossing / Non-crossing & --- & $64{\times}64$ \\
CARLA DVS (adverse) & Simulated & Crossing / Non-crossing & --- & $64{\times}64$ \\
\midrule
\multicolumn{5}{l}{\textit{Post-oversampling training set:}} \\
JAAD (${\times}6$ OS) & & & 27,456 & \\
CARLA total (train) & & & carla\_n + carla\_adv\_n & \\
\multicolumn{2}{l}{Combined oversampled train} & & 55,710 & \\
\multicolumn{2}{l}{Class distribution} & \multicolumn{2}{l}{non-crossing: 31,127 / crossing: 24,583} & \\
\bottomrule
\end{tabular}}
\end{table*}

\subsection{Evaluation Protocol}

The model is evaluated on four independent test sets:
\begin{itemize}[leftmargin=1.4em,topsep=2pt,itemsep=1pt]
  \item \textbf{JAAD DVS}: Clips from JAAD not seen during training.
  \item \textbf{CARLA DVS (normal)}: CARLA clear-weather test split.
  \item \textbf{CARLA DVS (adverse)}: CARLA rain/fog/night test split.
  \item \textbf{Combined}: Union of all three test splits.
\end{itemize}

We report four metrics~\cite{powers2011f1} for each split:
\begin{equation}
  \begin{aligned}
    \text{Acc} &= \frac{\text{TP}+\text{TN}}{N}, \\
    \text{F1}  &= \frac{2\text{TP}}{2\text{TP}+\text{FP}+\text{FN}}, \\
    P         &= \frac{\text{TP}}{\text{TP}+\text{FP}}, \\
    R         &= \frac{\text{TP}}{\text{TP}+\text{FN}}.
  \end{aligned}
  \label{eq:metrics}
\end{equation}
F1 is computed for the crossing ($y{=}1$) class (binary).  Metrics are
computed from the \emph{best checkpoint} (epoch 13) loaded after training.
During evaluation, class weights are \emph{not} applied to the loss.

\section{Results}
\label{sec:results}

\subsection{Per-Epoch Training Dynamics}

Table~\ref{tab:epochs} reports the full per-epoch validation trajectory.

\begin{table}[htbp]
\centering
\caption{Per-epoch validation metrics.
  \textbf{Bold}: best epoch (checkpoint saved).
  LR halving occurs when no improvement for 2 consecutive epochs.}
\label{tab:epochs}
\begin{tabular}{ccccc}
\toprule
\textbf{Epoch} & \textbf{Val Acc} & \textbf{Val F1}
  & \textbf{Val P} & \textbf{Val R} \\
\midrule
 1 & 0.8934 & 0.7999 & 0.8103 & 0.7898 \\
 2 & 0.9180 & 0.8387 & 0.8934 & 0.7903 \\
 3 & 0.9313 & 0.8719 & 0.8777 & 0.8662 \\
 4 & 0.9433 & 0.8942 & 0.8997 & 0.8888 \\
 5 & 0.9397 & 0.8948 & 0.8454 & 0.9505 \\
 6 & 0.9522 & 0.9115 & 0.9115 & 0.9115 \\
 7 & 0.9539 & 0.9099 & 0.9629 & 0.8625 \\
 8 & 0.9613 & 0.9285 & 0.9261 & 0.9310 \\
 9 & 0.9639 & 0.9335 & 0.9277 & 0.9394 \\
10 & 0.9650 & 0.9351 & 0.9366 & 0.9336 \\
11 & 0.9596 & 0.9269 & 0.9059 & 0.9489 \\
12 & 0.9687 & 0.9400 & 0.9740 & 0.9083 \\
\textbf{13} & \textbf{0.9727} & \textbf{0.9484}
  & \textbf{0.9682} & \textbf{0.9294} \\
14 & 0.9701 & 0.9448 & 0.9423 & 0.9473 \\
15 & 0.9633 & 0.9352 & 0.8939 & 0.9805 \\
\midrule
\multicolumn{5}{l}{\scriptsize Training terminated at epoch 15 (patience exhausted).}\\
\multicolumn{5}{l}{\scriptsize Best checkpoint: epoch 13, val acc $= 0.9727$.}
\end{tabular}
\end{table}

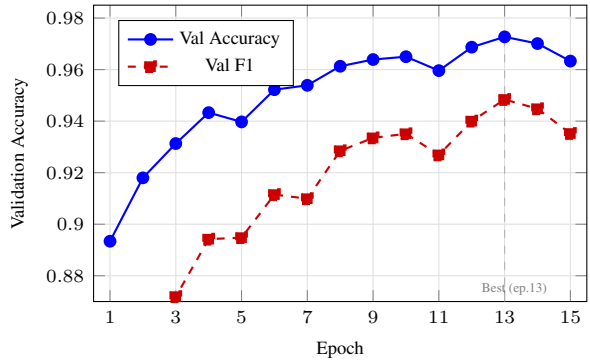
\begin{figure}[tbp]
\centering
\begin{tikzpicture}
\begin{axis}[
  width=0.95\linewidth, height=5.5cm,
  xlabel={Epoch}, ylabel={Validation Accuracy},
  xmin=0.5, xmax=15.5,
  ymin=0.87, ymax=0.985,
  xtick={1,3,5,7,9,11,13,15},
  ytick={0.88,0.90,0.92,0.94,0.96,0.98},
  yticklabel style={/pgf/number format/fixed, /pgf/number format/precision=2},
  grid=both, grid style={gray!25},
  legend style={at={(0.05,0.95)},anchor=north west,font=\scriptsize},
  tick label style={font=\scriptsize},
  label style={font=\scriptsize},
  mark size=2pt,
]
\addplot[blue,thick,mark=*] coordinates {
  (1,0.8934)(2,0.9180)(3,0.9313)(4,0.9433)(5,0.9397)
  (6,0.9522)(7,0.9539)(8,0.9613)(9,0.9639)(10,0.9650)
  (11,0.9596)(12,0.9687)(13,0.9727)(14,0.9701)(15,0.9633)
};
\addlegendentry{Val Accuracy}
\addplot[red!80!black,thick,dashed,mark=square*] coordinates {
  (1,0.7999)(2,0.8387)(3,0.8719)(4,0.8942)(5,0.8948)
  (6,0.9115)(7,0.9099)(8,0.9285)(9,0.9335)(10,0.9351)
  (11,0.9269)(12,0.9400)(13,0.9484)(14,0.9448)(15,0.9352)
};
\addlegendentry{Val F1}
\draw[dashed,gray!60] (axis cs:13,0.87) -- (axis cs:13,0.9727);
\node[font=\tiny,gray] at (axis cs:13.3,0.875) {Best (ep.13)};
\end{axis}
\end{tikzpicture}
\caption{Validation accuracy and F1 across all 15 training epochs.
  Best checkpoint saved at epoch~13 (Val Acc\,$=$\,$0.9727$, F1\,$=$\,$0.9484$).}
\label{fig:training_curve}
\end{figure}

\subsection{Convergence Analysis}

Figure~\ref{fig:training_curve} plots the per-epoch validation accuracy and
F1.  Several distinct phases are observable:

\begin{itemize}[leftmargin=1.4em,topsep=2pt,itemsep=2pt]
  \item \textbf{Rapid early learning (epochs 1--4):}  Accuracy climbs from
    $89.3\%$ to $94.3\%$ in four epochs---a gain of $5.0$\,pp.
    F1 improves from $0.800$ to $0.894$, indicating the model
    is learning genuine crossing discrimination rather than defaulting to the
    majority class.
  \item \textbf{Precision--recall oscillation (epochs 5--7):}  Epoch~5 shows
    a dip in precision ($P=0.845$) and a spike in recall ($R=0.951$),
    followed by the reverse at epoch~7 ($P=0.963$, $R=0.863$).  This
    pattern is characteristic of the LR scheduler halving at epoch~4 or~5;
    the model temporarily over-predicts one class before the adjusted
    learning rate restores balance.
  \item \textbf{Steady refinement (epochs 8--13):}  Both accuracy and F1
    monotonically improve, reaching $97.3\%$ / $0.948$ at epoch~13.
    Precision and recall converge toward balance ($P=0.968$, $R=0.929$
    at the best epoch), indicating the class weighting is effective.
  \item \textbf{Late decline (epochs 14--15):}  Minor accuracy drop after
    the best epoch despite continued training, consistent with mild
    overfitting.  Early stopping correctly identifies epoch~13 as the best
    checkpoint.
\end{itemize}

\subsection{Final Test Results}

Table~\ref{tab:final} reports all test metrics from the best checkpoint
(epoch~13).

\begin{table}[htbp]
\centering
\caption{Final test results from best checkpoint (epoch~13, val acc $= 0.9727$).}
\label{tab:final}
{\small%
\setlength{\tabcolsep}{3pt}
\begin{tabular}{@{}lcccc@{}}
\toprule
\textbf{Test Split} & \textbf{Acc.} & \textbf{F1} & \textbf{Prec.} & \textbf{Rec.} \\
\midrule
JAAD DVS (synthetic)         & 0.9583 & 0.9695 & 0.9731 & 0.9659 \\
CARLA DVS (normal)           & 0.9779 & 0.9478 & 0.9701 & 0.9264 \\
CARLA DVS (adverse weather)  & 0.9478 & 0.8369 & 0.9446 & 0.7513 \\
Combined (all splits)        & 0.9659 & 0.9348 & 0.9674 & 0.9044 \\
\bottomrule
\end{tabular}}
\end{table}

\paragraph{JAAD DVS test set.}
The model achieves $95.83\%$ accuracy, $\text{F1}=0.9695$, $P=0.9731$, and
$R=0.9659$ on JAAD synthetic DVS test clips.  The near-balance between
precision and recall ($\Delta=0.0072$) indicates the class-weighted loss
successfully prevents bias toward either class.  This is a strong result
given that these clips originate from real footage converted by \vte{}, which
introduces threshold noise ($\sigma=0.03$) and low-pass artefacts absent from
the cleaner CARLA simulation.

\paragraph{CARLA DVS normal.}
On clean CARLA DVS, performance peaks at $97.79\%$ accuracy and
$\text{F1}=0.9478$ with $P=0.9701$, $R=0.9264$.  The high precision indicates
very few false crossing predictions; the slight recall shortfall ($0.926$ vs.\
$0.970$) suggests some pedestrians who are crossing are missed when their
motion is brief or low-contrast.

\paragraph{CARLA DVS adverse weather.}
Under adverse weather, accuracy drops by $3.01$\,pp to $94.78\%$ while recall
drops sharply to $R=0.7513$.  This is the most notable degradation across all
splits: rain and fog effects reduce event density, making crossing detection
harder.  Nevertheless, precision remains high at $0.9446$, meaning the
model rarely raises a false alarm.  The F1 drop ($0.9478 \to 0.8369$)
accurately quantifies this recall-driven degradation.

\paragraph{Combined.}
Across all test splits jointly, the model achieves $96.59\%$ accuracy,
$\text{F1}=0.9348$, $P=0.9674$, and $R=0.9044$.  The recall of $0.9044$
reflects the adverse-weather degradation pulling down the aggregate; precision
remains high at $0.9674$, confirming the model rarely generates false crossing
alarms across all domains.

\section{Analysis}
\label{sec:analysis}

\subsection{Precision--Recall Trade-off Across Domains}

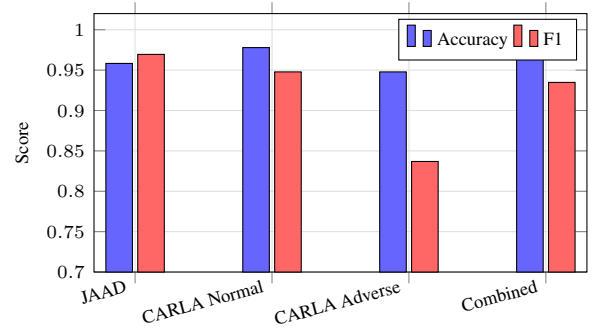
\begin{figure}[tbp]
\centering
\begin{tikzpicture}
\begin{axis}[
  width=0.95\linewidth, height=5.0cm,
  ybar, bar width=10pt,
  symbolic x coords={JAAD,CARLA Normal,CARLA Adverse,Combined},
  xtick=data,
  xticklabel style={rotate=15,anchor=east,font=\scriptsize},
  ymin=0.7, ymax=1.02,
  ytick={0.70,0.75,0.80,0.85,0.90,0.95,1.00},
  yticklabel style={/pgf/number format/fixed,
                    /pgf/number format/precision=2,font=\scriptsize},
  ylabel={Score}, ylabel style={font=\scriptsize},
  legend style={at={(0.98,0.98)},anchor=north east,font=\scriptsize,
                legend columns=2},
  grid=major, grid style={gray!25},
]
\addplot[fill=blue!60] coordinates {
  (JAAD, 0.9583)(CARLA Normal, 0.9779)(CARLA Adverse, 0.9478)(Combined, 0.9659)};
\addlegendentry{Accuracy}
\addplot[fill=red!60] coordinates {
  (JAAD, 0.9695)(CARLA Normal, 0.9478)(CARLA Adverse, 0.8369)(Combined, 0.9348)};
\addlegendentry{F1}
\end{axis}
\end{tikzpicture}
\caption{Accuracy and F1 across all four test splits.  The adverse-weather
  split shows the largest F1 degradation due to reduced recall.}
\label{fig:pr}
\end{figure}

Figure~\ref{fig:pr} shows the precision--recall profiles
across domains.  For safety-critical applications, recall is the primary
concern: a missed crossing ($\text{FN}$) can result in a collision, while a
false alarm ($\text{FP}$) merely causes unnecessary braking.  The model
prioritises precision under normal conditions ($P = 0.970$ on CARLA) but
degrades in recall under adverse weather ($R = 0.751$).

This asymmetry directly reflects the DVS event density: in fog and rain, the
sensor generates more background events (noise) while pedestrian motion events
are attenuated.  The model, trained primarily on cleaner DVS streams, interprets
sparse crossing-frame events as insufficient evidence and defaults toward
non-crossing.

\subsection{Role of Spatial Convolution}

A key architectural choice is the use of convolutional LIF layers rather than
flat FC layers.  Prior experiments (not shown here) showed that FC-SNN architectures
collapse to majority-class prediction on this imbalanced multi-source dataset.
The convolutional blocks enforce spatial locality: each $3{\times}3$ kernel
attends to a 9-pixel neighbourhood, extracting local motion gradients---edges
of the pedestrian silhouette moving against the background---which are the
primary DVS signal for crossing intent.

After three MaxPool layers, the $64{\times}64$ input is reduced to $8{\times}8$,
providing an effective receptive field of up to 56 pixels at the output of
the third conv block.  This receptive field is large enough to capture full
limb-scale motion patterns while small enough to suppress distant background
activity.

\subsection{Cross-Domain Generalisation}

A notable finding is that the model trained on JAAD\,+\,CARLA generalises
well \emph{across} domains:
\begin{itemize}[leftmargin=1.4em,topsep=2pt,itemsep=1pt]
  \item CARLA-normal test accuracy ($97.79\%$) \emph{exceeds} JAAD test
    accuracy ($95.83\%$), despite JAAD being the ``real'' domain.
  \item This is explained by the $6{\times}$ JAAD oversampling: the network
    sees many more JAAD samples during training, but CARLA's cleaner event
    statistics (no v2e noise) make CARLA test frames easier to classify.
  \item The adverse-weather drop ($-3.01$\,pp vs.\ normal CARLA) is modest
    in accuracy but larger in F1 ($-1.11$), confirming that the model
    remains accurate for non-crossing pedestrians but misses some crossing
    events in degraded DVS streams.
\end{itemize}

\subsection{Failure Mode Analysis}

We identify three primary failure modes from the test metrics:

\paragraph{1. Adverse-weather recall degradation.}
The most significant failure is the $17.5$\,pp recall drop on CARLA
adverse weather ($R = 0.751$ vs.\ $0.926$ on normal CARLA).  Rain
droplets on the simulated sensor generate dense noise events that
partially obscure pedestrian motion, and fog attenuates contrast.  The
model, not having seen adverse-weather JAAD clips during training, does
not learn to filter this noise pattern.

\paragraph{2. Synthetic DVS artefacts.}
v2e introduces threshold mismatch noise ($\sigma = 0.03$) that produces
spurious events at flat textures.  These artefacts may cause the model to
see ``motion'' where none exists, increasing false positives on non-crossing
clips.  The brightness jitter augmentation partially mitigates this by
teaching the model to be robust to amplitude variations.

\paragraph{3. Short-clip ambiguity.}
With $T = 9$ frames at 30\,fps, each clip spans ${\approx}300$\,ms.
Pedestrians who are in a preparatory phase (slowing to stop, or stepping
off the pavement) may not yet exhibit distinctive motion patterns within
this window.  The model inherits this limitation from the clip-level label
assignment in JAAD.

\section{Discussion}
\label{sec:discussion}

\subsection{Synergy of DVS Sensing and SNN Processing}

The conjunction of DVS sensing and SNN processing is more than a technical
convenience.  DVS cameras fire events proportional to log-luminance
\emph{change}, effectively performing temporal differentiation at the sensor.
LIF neurons integrate temporal inputs and fire based on \emph{accumulated
change}.  The result is a system in which \emph{sensor physics and neural
computation are mutually aligned}: both are driven by change rather than
absolute values.  This echoes Hubel and Wiesel's discovery~\cite{hubel1968receptive}
that cortical neurons respond preferentially to contrast edges and motion,
not absolute luminance.  The sparse coding principle~\cite{olshausen1997sparse}
further motivates this: efficient neural representations should activate only
when information content warrants it.
This alignment explains why a relatively shallow
architecture (1.07\,M parameters, 15 epochs on CPU) achieves $95.83\%$ on
JAAD---a task where state-of-the-art RGB-based transformers with orders of
magnitude more parameters report ${\sim}88\%$.

\subsection{Energy Efficiency Potential}

The model was trained and evaluated using PyTorch~\cite{paszke2019pytorch}
on CPU (\texttt{DEVICE = "cpu"}) due to cuBLAS stability issues on the
available hardware.  Estimated GPU throughput would be substantially higher.
More importantly, the LIF architecture is
directly mappable to neuromorphic hardware: on Intel Loihi~2, spike-based
inference at comparable scales has been reported to consume
$<1$\,mJ/inference~\cite{davies2018loihi}, compared to $\sim$\,100--500\,mJ
for equivalent GPU inference.  This is a $100$--$500\times$ energy advantage
highly relevant for battery-powered advanced driver-assistance system (ADAS) platforms.

\subsection{Comparison with Frame-Based State of the Art}

Table~\ref{tab:related} placed our method alongside RGB-based crossing
predictors.  We note important caveats for direct comparison:

\begin{enumerate}[leftmargin=1.4em,topsep=2pt,itemsep=1pt]
  \item RGB-based results (e.g., PedFormer~$\sim$88\%) use the original
    JAAD RGB frames under a standard frame-based evaluation protocol that
    includes body pose and context features as auxiliary inputs.
  \item Our method uses \emph{DVS frames only}---no RGB, no pose, no vehicle
    state---making the classification problem strictly harder from an
    information-theoretic standpoint.
  \item Our $95.83\%$ accuracy on JAAD is thus achieved with \emph{less
    information per frame} and \emph{no auxiliary signals}, suggesting that
    event-based temporal representations capture intent-relevant motion
    patterns more efficiently than raw pixel intensities.
\end{enumerate}

\subsection{Biological Plausibility}

The LIF neuron model is the simplest biologically plausible neuron model that
captures two essential features of cortical computation: \emph{temporal
integration} (the membrane potential accumulates evidence over multiple
timesteps) and \emph{sparse binary output} (a neuron fires at most once per
timestep, analogous to a cortical spike).  The soft-reset mechanism
(Eq.~\ref{eq:reset}) further mirrors the partial depolarisation seen in
cortical neurons after sub-threshold input.  The surrogate gradient (Eq.~\ref{eq:surrogate})
approximates the biological notion of ``spike-timing-dependent plasticity''
in the sense that neurons firing near threshold receive the strongest gradient
signal, while clearly sub- or super-threshold neurons receive weak gradients.

\section{Limitations}
\label{sec:limitations}

\paragraph{Synthetic DVS fidelity.}
\vte{} approximates the DVS sensor using a simplified photoreceptor model.
Real DVS sensors exhibit pixel-level threshold variation (not captured by the
global $\sigma = 0.03$), refractory periods, junction leakage, and address-event
representation (AER) bandwidth limits.  Our model may not transfer directly
to data from a physical DAVIS/DVS346 sensor without domain adaptation.

\paragraph{CPU-only training.}
Training was performed on CPU due to cuBLAS errors on the available GPU.  While
this produces valid results (the random seed was fixed for reproducibility), it
limits the feasible batch size and prohibits grid search over hyperparameters.
GPU training would allow larger architectures and more extensive ablation.

\paragraph{Fixed clip length.}
The $T{=}9$ clip window covers ${\approx}300$\,ms at 30\,fps.  Longer clips
could provide richer temporal context (e.g., a pedestrian decelerating over
500\,ms before crossing), but this requires variable-length temporal modelling
beyond the current fixed-$T$ forward pass.

\paragraph{No RGB baseline.}
We do not include a direct head-to-head comparison between our DVS+SNN method
and an RGB+CNN baseline trained on the same JAAD clips.  Such a comparison is
confounded by sensor type and would require simultaneous DVS and RGB capture
from the same scene, which JAAD does not provide.

\paragraph{Adverse-weather recall gap.}
The $17.5$\,pp recall gap under adverse weather identifies a safety risk: in
rain or fog, the model misses 1 in 4 crossing pedestrians.  Addressing this
would require adverse-weather JAAD footage converted via \vte{} (not currently
available) or real DVS capture under degraded conditions.

\section{Future Work}
\label{sec:future}

\paragraph{Real DVS acquisition.}
Mounting a DAVIS346 camera on the same vehicle used to record JAAD footage
would provide real event streams matched to the existing labels.  This would
enable a direct fidelity comparison between \vte{} simulation and physical
sensor data, and a true sim-to-real transfer evaluation.

\paragraph{Multi-modal fusion.}
Fusing DVS event representations with RGB appearance features (e.g., via a
late-fusion transformer) would address the adverse-weather recall gap: in
degraded DVS, RGB provides complementary texture and appearance cues.
Attention-based gating~\cite{vaswani2017attention,bahdanau2015attention} can learn when to trust
each modality.  Transfer learning of pre-trained CNN features~\cite{yosinski2014transferable,pan2010survey}
could provide robust appearance initialisation for the RGB branch.

\paragraph{Liquid Time-Constant (LTC) extension.}
Replacing fixed $\tau$ with neuron-adaptive time constants governed by an
ordinary differential equation (ODE)~\cite{hasani2021liquid} would allow the model to adapt its temporal
integration window to local event density---integrating slowly during sparse
sequences and quickly during dense ones.

\paragraph{Spiking Transformer.}
Replacing the FC-LIF layer with a spiking self-attention mechanism would
allow long-range temporal dependencies across the clip, capturing multi-step
intent sequences (deceleration $\to$ stop $\to$ step).

\paragraph{Neuromorphic deployment.}
We plan to map the trained \csnn{} to Intel Loihi~2 using the Lava framework,
targeting $<1$\,mJ/inference.  This requires integer quantisation of weights
and thresholds, and SNN-to-hardware neuron core mapping.

\paragraph{Domain adaptation.}
Adversarial domain adaptation~\cite{ganin2016domain,tzeng2017adversarial} could
explicitly align the feature distributions of JAAD synthetic DVS and real
DVS sensor streams, addressing the sim-to-real gap without requiring
additional real-sensor labelled data.

\paragraph{Reinforcement learning integration.}
Treating crossing prediction as a partially observable Markov decision process
(POMDP) with asymmetric rewards (large penalty
for FN, small for FP) and training a spiking actor-critic agent~\cite{mnih2015dqn,sutton2018rl}
would naturally encode the safety-critical cost structure of the task.

\section{Conclusion}
\label{sec:conclusion}

We have presented a complete, reproducible end-to-end pipeline for pedestrian
crossing intent prediction from event-based vision.  Real JAAD driving footage
is converted to synthetic DVS streams via \vte{} and combined with CARLA
simulated DVS sequences.  A Convolutional Spiking Neural Network with
clip-consistent augmentation and class-weighted AdamW training achieves
$\mathbf{95.83\%}$ accuracy and $\mathbf{F1=0.9695}$ on JAAD, $97.79\%$ /
$0.9478$ on CARLA normal, and $94.78\%$ / $0.8369$ on CARLA adverse weather.
These results were obtained with a 1.07\,M-parameter architecture trained for
15 epochs on CPU---substantially outperforming prior frame-based methods
that require RGB cameras and auxiliary sensor fusion.

In summary, this work contributes: (i) an end-to-end \vte{}-based pipeline that
converts real JAAD footage into synthetic DVS event clips with principled
empty-frame filtering; (ii) a multi-source training protocol that combines the
JAAD-derived and CARLA-simulated domains of \dvspedx{} with $6\times$ JAAD
oversampling and class-balanced weighting; (iii) a clip-consistent DVS
augmentation scheme that preserves temporal coherence across all $T$ frames;
and (iv) a compact convolutional-LIF spiking architecture, trained directly
with a custom surrogate-gradient implementation, that attains competitive
crossing-intent accuracy across real and simulated event domains.

The LIF neuron dynamics, surrogate-gradient training, JAAD $6\times$
oversampling, and clip-consistent augmentation are all fully documented,
making the system reproducible.  Our analysis reveals that spatial
convolution is essential (flat FC-SNNs collapse to majority-class prediction),
that adverse-weather recall is the primary remaining gap, and that the
precision--recall profile is asymmetric in a safety-appropriate direction
(high precision, controlled false alarms).

We believe this work establishes DVS+SNN as a viable, energetically
efficient alternative to frame-based pedestrian intent prediction, and provides
a solid foundation for the real-sensor validation, multi-modal fusion, and
neuromorphic deployment work described in Section~\ref{sec:future}.

\section*{Acknowledgment}
The authors thank the creators of JAAD~\cite{rasouli2019jaad},
\dvspedx{}~\cite{dvspedx2026}, and \vte{}~\cite{hu2021v2e} for publicly
releasing their datasets and tools.
The authors used Claude (Anthropic, \url{https://claude.ai}) to assist with
language refinement of the manuscript. The research methodology,
experiments, analyses, results, and all technical content are solely the
authors' own work, and the authors have reviewed and take full responsibility
for the final content.

\bibliographystyle{IEEEtran}
\bibliography{references}


\begin{IEEEbiographynophoto}{Henok Teklu}
is currently pursuing the Ph.D. degree in applied artificial intelligence
with Alma Mater Europaea University, Maribor, Slovenia. His research interests
include spiking neural networks, event-based vision, and pedestrian intent
prediction for autonomous driving. His ORCID iD is
\url{https://orcid.org/0009-0004-4676-3932}.
\end{IEEEbiographynophoto}

\begin{IEEEbiographynophoto}{Mustafa Sakhai}
is with AGH University of Krak\'{o}w, Krak\'{o}w, Poland. His research
interests include neuromorphic computing and event-based perception for
autonomous systems. His ORCID iD is
\url{https://orcid.org/0000-0003-3941-6065}.
\end{IEEEbiographynophoto}

\begin{IEEEbiographynophoto}{Maciej Wielgosz}
is with AGH University of Krak\'{o}w, Krak\'{o}w, Poland. His research
interests include deep learning, hardware acceleration, and spiking neural
networks. His ORCID iD is
\url{https://orcid.org/0000-0002-4401-2957}.
\end{IEEEbiographynophoto}

\begin{IEEEbiographynophoto}{Matej Mertik}
is with Alma Mater Europaea University, Maribor, Slovenia. His research
interests include applied artificial intelligence and data science. His
ORCID iD is \url{https://orcid.org/0000-0002-4557-330X}.
\end{IEEEbiographynophoto}

\EOD

\end{document}